\pdfoutput=1

\documentclass[11pt]{article}

\usepackage[]{ACL2023}

\usepackage{times}
\usepackage{latexsym}
\usepackage{times}
\usepackage{latexsym}
\usepackage{csquotes}
\usepackage{url}
\usepackage{tikz}
\usetikzlibrary{trees}
\usepackage{tikz-dependency}
\usepackage{tikz-qtree}
\usepackage{graphicx}
\usepackage{multicol,array,booktabs}
\usepackage{xcolor}
\usepackage{subcaption}
\usepackage{caption}
\usepackage{lipsum}
\usepackage{lipsum}
\usepackage{enumerate}
\usepackage{multirow}
\usepackage{colortbl}

\usepackage[T1]{fontenc}

\usepackage[utf8]{inputenc}
\usepackage{hyperref}

\usepackage{microtype}

\usepackage{inconsolata}

\title{Assessment of Pre-Trained Models Across Languages and Grammars}

\author
{
Alberto Muñoz-Ortiz, David Vilares and Carlos Gómez-Rodríguez\\
Universidade da Coru\~{n}a, CITIC \\
Departamento de Ciencias de la Computación y Tecnologías de la Información \\
Campus de Elvi\~{n}a s/n, 15071 \\ A Coru\~{n}a, Spain \\
\texttt{\{alberto.munoz.ortiz, david.vilares, carlos.gomez\}@udc.es} \\
}

\begin{document}
\maketitle
\begin{abstract}

We present an approach for assessing  how multilingual  large language models (LLMs) learn syntax in terms of multi-formalism syntactic structures. We aim to recover constituent and dependency structures by casting parsing as sequence labeling. To do so, we select a few LLMs and study them on 13 diverse UD treebanks for dependency parsing and 10 treebanks for constituent parsing. Our results show that: (i) the framework is consistent across encodings, (ii) pre-trained word vectors do not favor constituency representations of syntax over dependencies, (iii) sub-word tokenization is needed to represent syntax, in contrast to character-based models, and (iv) occurrence of a language in the pretraining data is more important than the amount of task data when recovering syntax from the word vectors.

\end{abstract}

\section{Introduction}
Large Language Models (LLMs) are the backbone for most NLP architectures. Their performance has not yet reached a plateau, and factors such as scale, language objective,  token segmentation or amount of pre-training time - among many others - play a role in their capabilities. 

To shed light on what is being learned, work on interpretability explains what these models encode in their representational space. Authors have explored whether these models exhibit stereotypical biases \cite{nadeem-etal-2021-stereoset}, encode facts \cite{poerner-etal-2020-e} or capture structural knowledge in multi-modal environments \cite{milewski-etal-2022-finding}.
Whether LLMs encode syntaxin their latent space has also been studied. In this respect, different \emph{probing frameworks}~\citep{KulmizevSchrodingers,belinkov-2022-probing} have been introduced to measure the syntactic capability of models, although authors such as \citet{hall-maudslay-cotterell-2021-syntactic} point out that we need to take this concept with caution, since they might not be completely isolating syntax.

Still, interpretability work on parsing focuses on either multilingual and mono-paradigm setups, or English and multi-paradigm setups. But we are not aware of \emph{multi-dimensional} work. This relates to the problem of square one bias in NLP research \cite{ruder-etal-2022-square}, that states that most work expands the current knowledge along just one dimension (e.g., a single language, or a single task). Related to our work, \citet{kulmizev-etal-2020-neural} study if LLMs showed preferences across two annotation \emph{styles}: deep syntactic and surface-syntactic universal dependencies, but both schemes were dependency-based. \citet{vilares2020parsing} did study two different syntactic formalisms, dependencies and constituents, and used a sequence-labeling-like recovery framework, relying on the pretraining architectures to associate output vectors with syntactic labels. We will build on top of this framework. Yet, they only studied English, and their analysis focused on static vectors and early LLMs; apart from other limitations that we discuss later.

\paragraph{Contribution} We move from square one bias in syntax assessment, and propose the first multi-paradigm, multilingual,
recovery framework for dependency and constituent structures learned by LLMs. We select representative LLMs that vary in scale, language pretraining objectives, and token representation formats. We then study their capability to retrieve syntax information from the pretrained representations on a diverse set of constituent and dependency treebanks, that vary in factors such as language family or size, as well as the presence or absence of their languages among the pretraining data of the LLMs. The code is available at \url{https://github.com/amunozo/multilingual-assessment}.

\section{Related work}\label{sec:relatedwork}
There is a long-standing effort in the NLP community to model syntax, either as a final goal or as a way to model compositionality. Yet, the ways in which this has been pursued have evolved with time. 

\paragraph{Modeling syntax in the pre-neural times.} 
Learning grammars through corpus-based approaches \cite{marcus-etal-1993-building,collins-1996-new,charniak1997statistical,petrov-klein-2007-improved} has been the dominating approach in the last decades. However, early models required extensive feature engineering to obtain competitive parsers. This suggested that support vector machines (SVMs) had severe limitations understanding language structure, and needed the help of parsing algorithms \cite{nivre-2008-algorithms,martins-etal-2010-turbo}, language-dependent features \cite{ballesteros-nivre-2012-maltoptimizer-system}, or tree-kernels \cite{lin-etal-2014-descending,zhang-li-2009-tree} to model syntax properly.

\paragraph{Modeling syntax in neural times.} With the rise of word vectors \cite{mikolov2013distributed}, LSTMs~\cite{hochreiter1997long} and Transformers~\cite{NIPS2017_3f5ee243}, modeling structure has become less relevant to obtain a good performance, both for parsing and downstream tasks. For instance, while the classic parser by \citet{zhang-nivre-2011-transition} used a rich set of features (including third-order, distance, and valency features, among others) to be competitive, the parser by \citet{chen-manning-2014-fast} only needed 18 word and PoS tag features (and 6 dependency features) to obtain strong results, which was possible thanks to their reliance on pre-trained word vectors and neural networks. The need for feature engineering was reduced further with bidirectional LSTMs, e.g., \citet{kiperwasser-goldberg-2016-simple} showed that four vectors corresponding to elements in the buffer and the stack sufficed to obtain state-of-the-art performance, while \citet{shi-etal-2017-fast} showed that competitive accuracies were possible with only two features. 

\paragraph{Modeling syntax in the era of language models.} 
In the context of these (almost) end-to-end parsers performing very competitively without the need of explicitly modeling syntactic linguistic features, recent efforts have been dedicated to interpret to what extent syntax is encoded in the representational space of neural networks, and in particular of LLMs. \citet{tenney2019you} and \citet{liu-etal-2019-linguistic} proposed probing frameworks for partial parsing, in the sense that they tried to demonstrate that certain syntactic information, such as dependency types, was encoded in pre-trained models. \citet{vilares2020parsing} defined a probing framework for full dependency and constituent parsing. They cast dependency and constituent parsing as sequence labeling and associated output vectors with syntactic labels by freezing their models. \citet{hewitt-manning-2019-structural} proposed a structural probing framework and identified that pre-trained models encoded a linear transformation that indicates the distance between words in a dependency tree. The framework was later upgraded to extract directed and labeled trees, while using fewer parameters \cite{muller-eberstein-etal-2022-probing}. \citet{hewitt-liang-2019-designing} pointed out that we need to be careful with probing frameworks, since the probe might be learning the linguistic task itself, instead of demonstrating the presence of the target linguistic property. For that, they recommend to use control experiments, and relied on control tasks, i.e., learning a random task with the same dimensional output space. \citet{hall-maudslay-cotterell-2021-syntactic} showed that semantic cues in the data might guide the probe and therefore they might not isolate syntax, although their experiments still outperformed the baselines. \citet{muller-eberstein-etal-2022-sort} found the most suitable pre-trained LLMs to plug into a dependency parser for a given treebank. Particularly, they proposed to rank frozen encoder representations by determining the percentage of trees that are recoverable from them, and based on that ranking choose which LLM to plug. Focused on morphology, \citet{stanczak-etal-2022-neurons} showed that subsets of neurons model morphosyntax across a variety of languages in multilingual LLMs.

\section{Multilingual probing frameworks}
Let $w$ = $[w_1,w_2,...,w_n]$ be an input sentence. We are interested in linear probing frameworks that can associate a sequence of word vectors $\vec{w}$ = $[\vec{w}_1,\vec{w}_2,...,\vec{w}_n]$ to a given linguistic property $[p_1,p_2,...,p_n]$. For some properties, the mapping can be quite direct, such as for instance the case of part-of-speech (PoS) tagging (by putting a linear layer on top of $w$ and outputting the PoS tag category), or lexical semantics (e.g. computing word vector similarity). We want an analogous mapping, but for multiple syntactic formalisms. In this case, the association is not trivial since syntactic parsing is a tree-based structured prediction problem. Also, we are interested in multilingual pre-trained models, which have gained  interest in recent years. Then, the goal is to associate their word vectors to an estimate of to what extent characteristics of a given formalism are encoded in their representational space, and whether this can differ across dimensions such as tested models, formalisms, and treebanks.

\paragraph{Linear probing framework for parsing}
We take the study by \citet{vilares2020parsing} as our starting point. However, we first identify some weaknesses in their work: (i) it is limited to English, (ii) they do not give specific estimates of the amount of trees recoverable with respect to control experiments, and (iii) they only test one type of tree linearization. For the latter, the main motivation, in particular for the case of dependency parsing, was that the chosen linearization had performed the best in previous work \cite{strzyz-etal-2019-viable} when training from scratch a transducer without pre-training. However, later work suggests that that is debatable: for instance, \citet{munoz-ortiz-etal-2021-linearizations} show that different tree linearizations might be better suited to different languages, and \citet{vacareanu-etal-2020-parsing}'s results indicate that other encodings worked better when pre-trained language models are used.

\noindent To recover dependency and constituent structures, we will represent the trees
using existing encodings for parsing as sequence labeling \cite{gomez-rodriguez-vilares-2018-constituent,strzyz-etal-2019-viable}.
Under this configuration, the interaction between learning a model and searching for linguistic properties is now direct. We can use probing architectures that rely entirely on the pretrained representations, and simply add a linear layer on top to map continuous vectors to discrete labels. We can expect that the capabilities of the output layer are not enough to learn the syntactic tasks at hand by themselves, so it must rely on the quality of the pretrained representations. Yet, we also will include control baselines that we will discuss later.

\paragraph{Research questions} We want to answer two questions: 
(i) how much syntax is recoverable from different LLMs? and (ii) how is it affected by aspects such as the models, the type of formalism, and the pretraining and assessment data?

In what follows, we describe the sequence labeling encodings, 
both for dependency and constituent paradigms (\S \ref{ssec:parsing_sl}), and the specifics of the probing setup used for our experiments (\S \ref{ssec:setup}).

\subsection{Sequence labeling encodings of syntax}\label{ssec:parsing_sl}

Parsing as sequence labeling can be defined as learning a function $f_{n}: V^n \rightarrow L^n$ to map a sequence of words into a sequence of linearized labels that can be decoded to fully recover a constituent or dependency tree. Here we are not interested in the parsers \emph{per se}, but in whether the sequence-labeling encodings defined for them provide a simple, lossless representation of dependency and constituent trees that is useful for probing. In what follows, we briefly describe these representations.

\subsubsection{Dependency parsing}
Dependencies between tokens can be encoded using labels of the form $(x_i,l_i)$, where $x_i$ is a subset of the arcs related to the token $w_i$, and $l_i$ denotes the dependency relation \cite{strzyz-etal-2019-viable}. There are different ways of encoding $x_i$\footnote{To ensure that the labels produce a valid tree, we apply the postprocessing described in the paper of each encoding.}. We compare three families of linearizations (due to brevity, we refer to the references below for the details):
\begin{figure}[hbpt]
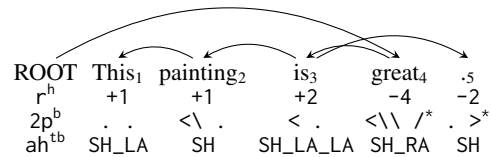

\begin{center}
\begin{dependency}[hide label, arc edge, arc angle=30, edge unit distance=1.2ex]
\small
    \begin{deptext}
        \small{ROOT}\&This\textsubscript{1}\& painting\textsubscript{2}\&is\textsubscript{3}\&great\textsubscript{4}\&.\textsubscript{5}\\
        \small{\texttt{r\textsuperscript{h}}}  \& \small{\texttt{+1}} \&\small{\texttt{+1}}  \& \small{\texttt{+2}} \& \small{\texttt{-4}} \& \small{\texttt{-2}} \\
        \small{\texttt{2p\textsuperscript{b}}} \& \small{\texttt{. .}} \& \small{\texttt{<\symbol{92} .}}  \& \small{\texttt{< .}} \& \small{\texttt{<\symbol{92}\symbol{92} /\textsuperscript{*}}} \& \small{\texttt{. >\textsuperscript{*}}} \\
        \small{\texttt{ah\textsuperscript{tb}}} \& \small{\texttt{SH\_LA}} \&  \small{\texttt{SH}} \& \small{\texttt{SH\_LA\_LA}}  \& \small{\texttt{SH\_RA}}  \& \small{\texttt{SH}} \\
    \end{deptext}
    \depedge{3}{2}{}
    \depedge{4}{3}{}
    \depedge{5}{4}{}
    \depedge{1}{5}{}
    \depedge{4}{6}{}
\end{dependency}
\end{center}

\caption{\label{fig:dep_example} Example of a dependency tree linearization. Dependency types are omitted. For \texttt{2p\textsuperscript{b}}, the dot indicates no bracket in the first and/or second plane.}
\end{figure}

\paragraph{Head-selection} \cite{spoustova,li-etal-2018-seq2seq,strzyz-etal-2019-viable}. 
$x_i$ encodes the dependency arc pointing directly to $w_i$. This can be done using an absolute index or a relative offset computing the difference between $w_i$'s index and its head. 
We use (\texttt{r\textsuperscript{h}}) encoding where the head of $w_i$ is the $x_i$th word to the right, if $x_i>0$, and the $x_i$th word to the left if $x_i<0$.\footnote{There are other head-selection encodings where the offset depends on some word property, e.g., PoS tags like in \cite{vilares2020parsing}, but using these encodings can blur the probing, since we need to access such external information.}

\paragraph{Bracketing-based} \cite{yli-jyra-gomez-rodriguez-2017-generic,strzyz-etal-2020-bracketing}. 
$x_i$ encodes the arcs using strings of brackets to represent a subset of the incoming and outgoing arcs of $w_i$ and its direct neighbors. We use a 2-planar bracketing encoding (\texttt{2p\textsuperscript{b}}) that uses two independent planes of brackets to encode non-projective trees.

\paragraph{Transition-based} \cite{gomez-rodriguez-etal-2020-unifying}. 
$x_i$ encodes a sub-sequence of the transitions that are generated by a left-to-right transition-based parser. Given a transition list $t=t_1,...,t_m$ with $n$ read transitions, $t$ is split into $n$ sub-sequences such that the $i$th sub-sequence is assigned to $w_i$. We use a mapping from the arc-hybrid algorithm (\texttt{ah\textsuperscript{tb}}) \cite{kuhlmann-etal-2011-dynamic}. These mappings are implicit and often perform worse than more direct encodings, but they are learnable.
\medskip

\noindent These encodings produce labels with different information. Following Figure \ref{fig:dep_example}, for $w_2$ (painting), the $2p^b$ encoding states that the previous word $w_1$ has one incoming arc from the right ("$<$" symbol, but it does not say from where, as that information is encoded in other labels) and that $w_2$ has one outgoing arc to the left ("\symbol{92}" symbol, but it does not specify where). For the transition-based encoding,  the mapping is less straightforward across words, but still connected to them. For instance, for $w_1$ (`This') the label indicates that the $w_1$ has no connection to $w_0$, that it is a dependent of $w_1$, and that it has no children. The motivation to compare encodings is to test: (i) the consistency of the framework, i.e., if trends across LLMs remain, and (ii) to see what information is easier to recover when the LLM weights are frozen.

\subsubsection{Constituent parsing}
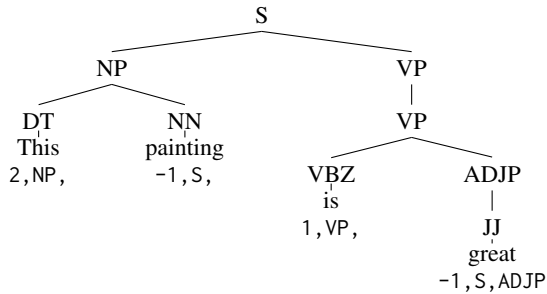
\begin{figure}[hbpt]
    \centering
    \small
    \setlength{\textwidth}{8cm}
    \setlength{\columnsep}{1cm}
    \begin{tikzpicture}[%
      sibling distance=.8cm,
      every node/.append style={align=center},
      level distance=20pt,
      scale=1,
      ]
    \Tree [.S 
    [.NP [.DT This\\\texttt{2,NP,} ] [.NN painting\\\texttt{-1,S,} ] ] 
    [.VP 
        [.VP [.VBZ is\\\texttt{1,VP,} ] 
            [.ADJP [.JJ great\\\texttt{-1,S,ADJP} ] ] 
        ] 
    ] 
]
    \end{tikzpicture}
    \caption{Example of a constituent tree linearization.}
    \label{fig:const_example}
\end{figure}

We here use the encoding approach by \citet{gomez-rodriguez-vilares-2018-constituent}, which encodes common levels in the tree between pairs of tokens.\footnote{To our knowledge, when we did the experiments, this encoding (together with variants) was the only available family of sequence-labeling encodings for constituency parsing. Contemporaneously to the end of this work, another family of encodings - based on the tetra-tagging \cite{kitaev-klein-2020-tetra} - has been proposed and implemented as a pure tagging approach \cite{amini-cotterell-2022-parsing}.} The labels are of the form $(n_i, c_i,u_i)$. The element $n_i$ encodes the number of tree levels that are common between $w_i$ and $w_{i+1}$, computed as the difference with respect to $n_{i-1}$. The element $c_i$ encodes the lowest non-terminal symbol that is shared between those two words. $u_i$ encodes the leaf unary branch located at $w_i$, if it exists. An example is shown in Figure \ref{fig:const_example}.

\subsection{Probing architecture}\label{ssec:setup}
We use a 1-layered feed-forward network on top of the LLMs to predict the labels. We propose three setups (training hyperparameters are detailed in Appendix \ref{app:hyperparameters}):
\paragraph{Frozen weights (\texttt{frz})} The LLM weights are frozen and only the weights of the linear output layer are updated during fine-tuning.
\paragraph{Random weights (\texttt{rnd})} Only the weights of the linear classifier layer are updated, but the weights of the encoders are randomized. We aim to prevent misleading conclusions in the hypothetical case that the linear layer can learn the mapping itself, i.e., we use this setup as a lower bound baseline. It is also a control experiment, as the difference between the results of this setup and the \texttt{frz} setup would be the measure we are looking for to estimate the amount of syntax information encoded in the representational space of pre-trained LLMs.

\paragraph{Fine-tuned weights (\texttt{ftd})} A fine-tuned LLM where  all weights are updated, i.e., this setup is used as an upper bound baseline.

\subsection{Multilingual Language Models}
The method here proposed is model-agnostic. Our aim is not to obtain the highest results or to use the largest LLM. We select a few LLMs that are representative and runnable with our resources:

\paragraph{\texttt{mBERT}}~\cite{devlin-etal-2019-bert} It uses WordPiece tokenization. While subword tokenizers are effective with representative splits, they yield suboptimal subtokens for low-resource languages \cite{agerri-etal-2020-give,virtanen2019multilingual},
as wrong subtokens will not encode meaningful information. mBERT is pretrained on 104 languages from the dump of the largest Wikipedias.

\paragraph{\texttt{xlm-roberta}}~\cite{conneau-etal-2020-unsupervised} A multi-lingual LLM trained as RoBERTa \cite{liu2019roberta}. It has the same architecture as BERT, but only pretrained on the masked word prediction task and uses a byte-level BPE for tokenization. 
It has been pretrained on 2.5TB of filtered CommonCrawl data that contains text in 100 languages (XLM-100), and for longer time than \texttt{mBERT}.

\paragraph{\texttt{canine (-c and -s)}} \cite{clark-etal-2022-canine} 
It uses char-based tokenization, which is believed to perform better in languages that are challenging for subword tokenization, such as those with vowel harmony. It eliminates the issue of unknown tokens. It is pre-trained on masked language modeling and next sentence prediction on the same data as \texttt{mBERT}: \texttt{canine-c} is pretrained using a char-level loss, while \texttt{canine-s} includes a previous subword tokenization to predict masked subword tokens.\\

\noindent In all models, labels are first broken down into subtokens before being processed by the LLMs to assign them to the $n$ input tokens. The classifier layer then assigns a label to each subtoken  (i.e. subword for \texttt{mBERT} and \texttt{xlm-roberta} and character for \texttt{canine}). Then, we select the label assigned to the first sub-element, which is a common approach.

\section{Methodology and Experiments}

\paragraph{Data for dependency parsing} For the assessment of dependency structures, we selected 13 Universal Dependencies \citep[UD 2.9;][]{nivre-etal-2020-universal} treebanks from different language families and with different amounts of annotated data. Although \texttt{mBERT}, \texttt{xlm-roberta}, and \texttt{canine} have been pre-trained on different (multilingual) crawled datasets, we select treebanks whose languages are either present in all our LLMs' pretraining data or in none of them (although presence proportions might vary in the case of \texttt{xlm-roberta}). For more details,  see Table \ref{tab:dep_languages}. Data sizes have been obtained from \citet{wu-dredze-2020-languages} for Wiki-100 and \citet{conneau-etal-2020-unsupervised} for XLM-100.

\begin{table}[hbpt!]
    \centering
    \scriptsize
    \addtolength{\tabcolsep}{-0.5pt}
    \begin{tabular}{llrrrr}
    \toprule
    \multirow{2}{*}{Treebank} & \multirow{2}{*}{Family} & \multirow{2}{*}{\# Trees} & \multirow{2}{*}{\# Tokens} & Wiki-100 & XLM-100 \\ 
    & & & & size (GB) & size (GB) \\
    \midrule
    Skolt Sami\textsubscript{Giellagas} & Sami & 200 & 2\,461 & - & -\\
    Guajajara\textsubscript{TuDeT} & Tupi-Guarani & 284 & 2\,052 & - & -\\
    Ligurian\textsubscript{GLT} & Romance & 316 & 6\,928 & - & -\\
    Bhojpuri\textsubscript{BHTB} & Indic & 357 & 6\,665 & - & - \\
    Kiche\textsubscript{IU} & Mayan & 1\,435 & 10\,013 & - & -\\
    Welsh\textsubscript{CCG} & Celtic & 2\,111 &  41\,208 & <0.1 & 0.8 \\
    Armenian\textsubscript{ArmTDP} & Armenian & 2\,502 & 52\,630 & 0.2-0.4 & 5.5\\
    Vietnamese\textsubscript{VTB} & Viet-Muon) & 3\,000 & 43\,754 & 0.4-0.7 & 137.3\\
    Chinese\textsubscript{GSDSimp} & Sinitic & 4\,997 & 128\,291 & 1.4-2.8 & 46.9\\
    Basque\textsubscript{BDT} & Basque & 8\,993 & 121\,443 & 0.1-0.2 & 2.0\\
    Turkish\textsubscript{BOUN} & Turkic & 9\,761 & 122\,383 & 0.4-0.7 & 20.9\\
    Bulgarian\textsubscript{BTB} & Slavic & 11\,138 & 146\,159 & 0.2-0.4 & 57.5 \\
    Ancient Greek\textsubscript{Perseus} & Greek & 13\,919 & 202\,989 & - & -\\
    \bottomrule
    \end{tabular}
    \caption{Dependency treebanks used in this work.}
    \label{tab:dep_languages}
\end{table}

\paragraph{Data for constituent parsing} We assess constituent structures on the PTB \cite{marcus-etal-1993-building}, the CTB \cite{xue2005penn}, and 8 constituent treebanks from the SPMRL shared task \citep{seddah-etal-2014-introducing}\footnote{
We do not have the license for the Arabic treebank.}, whose languages are shown in Table \ref{tab:const_languages}. 

\paragraph{Language disparity}
We use (mostly) different languages for each paradigm. For constituent treebanks, we only have access to rich-resource languages, so we prioritize diversity for dependencies. Comparing languages across syntax paradigms is not particularly useful, due to varying metrics, annotation complexity, and treebank comparisons. Instead, we compare error reductions against control models to estimate the recoverability of specific syntactic formalisms by an LLM (see \S \ref{sec:results}).

\begin{table}[hbpt!]
    \centering
    \scriptsize
    \begin{tabular}{llrrrr}
    \toprule
    \multirow{2}{*}{Treebank} & \multirow{2}{*}{Family} & \multirow{2}{*}{\# Trees} & \multirow{2}{*}{\# Tokens} & Wiki-100 & XLM-100 \\ 
    & & & & size (GB) & size (GB) \\ \midrule  
    Swedish & Germanic & 5\,000 & 81\,333  & 0.7-1.4 & 12.1\\ 
    Hebrew & Semitic & 5\,000 & 133\,047 & 0.4-0.7 & 31.6 \\
    Polish & Slavic & 6\,578 & 73\,357 & 1.4-2.8 & 44.6 \\
    Basque & Basque & 7\,577 & 103\,946 & 0.1-0.2 & 2.0 \\
    Hungarian & Finno-Ugric & 8\,146 & 178\,278 & 0.8-1.4 & 58.4 \\
    French & Romance & 14\,759 & 457\,873 & 2.8-5.7 & 56.8 \\
    Korean & Korean & 23\,010 & 319\,457 & 0.4-0.7 & 54.2\\
    English & Germanic & 39\,832 & 989\,861 & 11.3-22.6 & 300.8\\
    German & Germanic & 40\,472 & 760\,003 & 2.8-5.7 & 66.6\\
    Chinese & Sinitic & 50\,734 & 1\,235\,267 & 1.4-2.8 & 46.9\\     
    \bottomrule
    \end{tabular}
    \caption{Constituent treebanks used in this work}
    \label{tab:const_languages}
\end{table}

\paragraph{Metrics}
For dependency parsing, we use Labeled Attachment Score (LAS). For constituent parsing, we use the labeled bracketing F1-score.

\section{Results}\label{sec:results}
\definecolor{gray}{gray}{0.6}
\newcolumntype{g}{>{\color{gray}}c}
\begin{table*}[t!]
    \centering
    \scriptsize
    \begin{tabular}{l|gcgcgc|gcgcgc|gcgcgc|gcgcgc}
    \toprule
    \multirow{3}{*}{Treebank}& \multicolumn{6}{c|}{\texttt{mBERT}} & \multicolumn{6}{c|}{\texttt{xlm-roberta}} & \multicolumn{6}{c|}{\texttt{canine-c}} & \multicolumn{6}{c}{\texttt{canine-s}}\\
     & \multicolumn{2}{c}{\texttt{2p\textsuperscript{b}}} & \multicolumn{2}{c}{\texttt{ah\textsuperscript{tb}}} & \multicolumn{2}{c}{\texttt{r\textsuperscript{h}}}
     & \multicolumn{2}{c}{\texttt{2p\textsuperscript{b}}} & \multicolumn{2}{c}{\texttt{ah\textsuperscript{tb}}} & \multicolumn{2}{c}{\texttt{r\textsuperscript{h}}}
     & \multicolumn{2}{c}{\texttt{2p\textsuperscript{b}}} & \multicolumn{2}{c}{\texttt{ah\textsuperscript{tb}}} & \multicolumn{2}{c}{\texttt{r\textsuperscript{h}}}
     & \multicolumn{2}{c}{\texttt{2p\textsuperscript{b}}} & \multicolumn{2}{c}{\texttt{ah\textsuperscript{tb}}} & \multicolumn{2}{c}{\texttt{r\textsuperscript{h}}}\\
     & \texttt{rnd} & \texttt{frz} & \texttt{rnd} & \texttt{frz} & \texttt{rnd} & \texttt{frz} & \texttt{rnd} & \texttt{frz} 
     & \texttt{rnd} & \texttt{frz} & \texttt{rnd} & \texttt{frz} & \texttt{rnd} & \texttt{frz} & \texttt{rnd} & \texttt{frz} 
     & \texttt{rnd} & \texttt{frz} & \texttt{rnd} & \texttt{frz} & \texttt{rnd} & \texttt{frz} & \texttt{rnd} & \texttt{frz}\\
     \midrule
     \textit{Skolt Sami} & 11.5 & 9.2 & 8.0 & 8.4 & 10.4 & 13.5 & 14.2 & 6.9 & 6.5 & 3.0 & 10.5 & 8.5 & 7.6 & 7.2 & 9.0 & 5.1 & 9.2 & 6.2 & 10.5 & 10.3 & 9.3 & 8.0 & 9.2 & 8.0 \\
     \textit{Guajajara} & 31.8 & 30.9 & 26.4 & 26.4 & 27.9 & 22.2 & 35.3 & 19.0 & 26.4 & 12.1 & 31.1 & 12.2 & 29.2 & 22.2 & 24.0 & 15.3 & 27.9 & 22.2 & 29.4 & 29.8 & 22.0 & 21.5 & 27.9 & 27.9 \\
     \textit{Ligurian} & 2.9 & 7.2 & 12.1 & 21.7 & 16.6 & 21.2 & 3.8 & 1.6 & 14.6 & 9.8 & 16.7 & 6.6 & 4.4 & 3.6 & 12.8 & 8.2 & 13.2 & 10.5 & 4.8 & 5.7 & 10.8 & 11.7 & 13.5 & 12.5 \\
     \textit{Bhojpuri} & 14.4 & 17.0 & 17.3 & 26.0 & 24.3 & 28.3 & 13.2 & 11.8 & 17.5 & 18.6 & 24.6 & 26.8 & 13.3 & 9.1 & 16.9 & 4.8 & 22.4 & 13.8 & 13.4 & 11.3 & 17.7 & 10.3 & 22.4 & 17.8\\
     \textit{Kiche} & 45.2 & 51.0 & 43.0 & 49.0 & 42.3 & 45.6  & 41.6 & 33.4 & 41.2 & 31.5 & 39.5 & 33.1 & 47.4 & 25.7 & 43.2 & 24.3 & 43.1 & 25.3 & 47.8 & 43.1 & 43.3 & 40.7 & 43.1 & 40.3 \\
     Welsh & 22.4 & 44.9 & 23.0 & 42.6 & 22.3 & 43.6 & 23.6 & 28.0 & 23.3 & 29.7 & 21.6 & 30.1 & 25.7 & 12.5 & 24.6 & 11.9 & 27.9 & 15.6  & 25.8 & 20.5 & 24.3 & 19.8 & 27.9 & 23.9\\
     Armenian & 15.1 & 38.8 & 13.5 & 33.7 & 19.9 & 34.8 & 13.3 & 31.0 & 12.4 & 25.9 & 18.1 & 30.9 & 15.1 & 13.8 & 13.5 & 10.3 & 18.9 & 16.8 & 14.8 & 19.9 & 13.8 & 15.2 & 18.9 & 21.2 \\
     Vietnamese & 14.7 & 37.4 & 19.6 & 37.8 & 14.7 & 31.8 & 14.7 & 24.6 & 18.4 & 26.7 & 14.6 & 19.2 & 13.9 & 10.0 & 13.3 & 6.3 & 16.1 & 12.0 & 14.1 & 15.3 & 13.4 & 13.3 & 16.1 & 16.7\\ 
     Chinese & 11.0 & 42.1 & 14.2 & 39.1 & 21.0 & 38.8 & 1.9 & 17.6 & 5.7 & 18.9 & 11.5 & 25.1 & 13.6 & 15.6 & 15.0 & 14.7 & 20.1 & 19.4 & 13.6 & 24.9 & 15.0 & 23.5 & 20.4 & 27.3\\
     Basque & 17.9 & 45.5 & 16.3 & 41.9 & 19.6 & 40.2 & 17.2 & 40.9 & 15.6 & 37.6 & 18.7 & 32.8 & 18.8 & 14.8 & 16.2 & 12.6 & 20.7 & 16.4 & 18.8 & 22.4 & 16.2 & 19.5 & 20.7 & 22.9\\
     Turkish & 20.0 & 42.9 & 19.2 & 41.5 & 25.1 & 40.8 & 19.4 & 41.3 & 18.7 & 40.2 & 23.9 & 39.1 & 18.9 & 18.5 & 16.4 & 14.1 & 21.8 & 20.1 & 18.8 & 23.7 & 16.5 & 21.5 & 21.8 & 24.3\\
     Bulgarian & 20.8 & 63.4 & 22.4 & 56.4 & 25.7 & 54.3 & 22.3 & 55.3 & 22.7 & 47.9 & 26.2 & 46.0 & 23.9 & 18.0 & 21.4 & 16.2 & 26.3 & 21.2 & 23.8 & 29.4 & 21.2 & 25.5 & 26.3 & 30.5\\
     \textit{A. Greek} & 6.6 & 23.7 & 14.5 & 24.3 & 14.9 & 23.8 & 5.4 & 23.7 & 12.9 & 27.3 & 14.3 & 25.6 & 13.4 & 11.3 & 15.4 & 14.0 & 17.3 & 16.4 & 13.6 & 18.4 & 15.4 & 20.3 & 17.3 & 20.6\\
    \midrule
    Average & 18.0 & 34.9 & 19.2 & 34.5 & 22.2 & 34.2 & 17.4 & 25.8 & 18.1 & 25.3 & 20.9 & 25.8 & 18.9 & 13.9 & 18.6 & 12.1 & 21.9 & 16.6 & 19.2 & 21.1 & 18.4 & 19.3 & 22.0 & 22.6 \\
    \bottomrule
    \end{tabular}
    \caption{LAS for the test sets of the dependency treebanks. LLMs and dependency encodings analyzed for the \texttt{frz} and \texttt{rnd} setups. Languages \emph{in italics} are  absent among the crawled data used to pre-train the LLMs.}
    \label{tab:dep_results}
\end{table*}

We present the assessment for dependency structures in \S\ref{section-results-dep}, and for constituent structures in \S\ref{section-results-const}. 

\subsection{Dependency parsing results}\label{section-results-dep}

We break down the results comparing frozen vs: (i) random, and (ii) fine-tuned weights.

\paragraph{Frozen (\texttt{frz}) \emph{vs} random weights (\texttt{rnd}) setups} Table \ref{tab:dep_results} shows the LAS results across treebanks and dependency encodings (head-based, bracketing-based, and transition-based). For \texttt{mbert} and \texttt{xlm-roberta} the performance in the \texttt{frz} setup clearly surpasses the \texttt{rnd} baseline, i.e., the control experiment. The results suggest that under the frozen setups, \texttt{mbert} is better than \texttt{xlm-roberta} at recovering dependencies, although pre-trained \texttt{xlm-roberta} models are usually better at downstream tasks \cite{liu2019roberta}.
The ranking of the LLMs is stable across treebanks. The LAS scores across encodings are in a similar range, and the average LAS across different encodings is very similar too (bottom row in Table \ref{tab:dep_results}). On the other hand, the results for \texttt{canine} do not surpass the lower bound baseline in most cases. This is unlikely to be because of a bad fitting, since the random weights baselines perform almost the same across pre-trained models, encodings and treebanks. Also, while \texttt{canine-s} outperforms
the random baseline for the highest-resourced languages, \texttt{canine-c} underperforms it for all languages except for Chinese.

\begin{figure}[hpbt]
    \centering
    \includegraphics[width=0.45\textwidth]{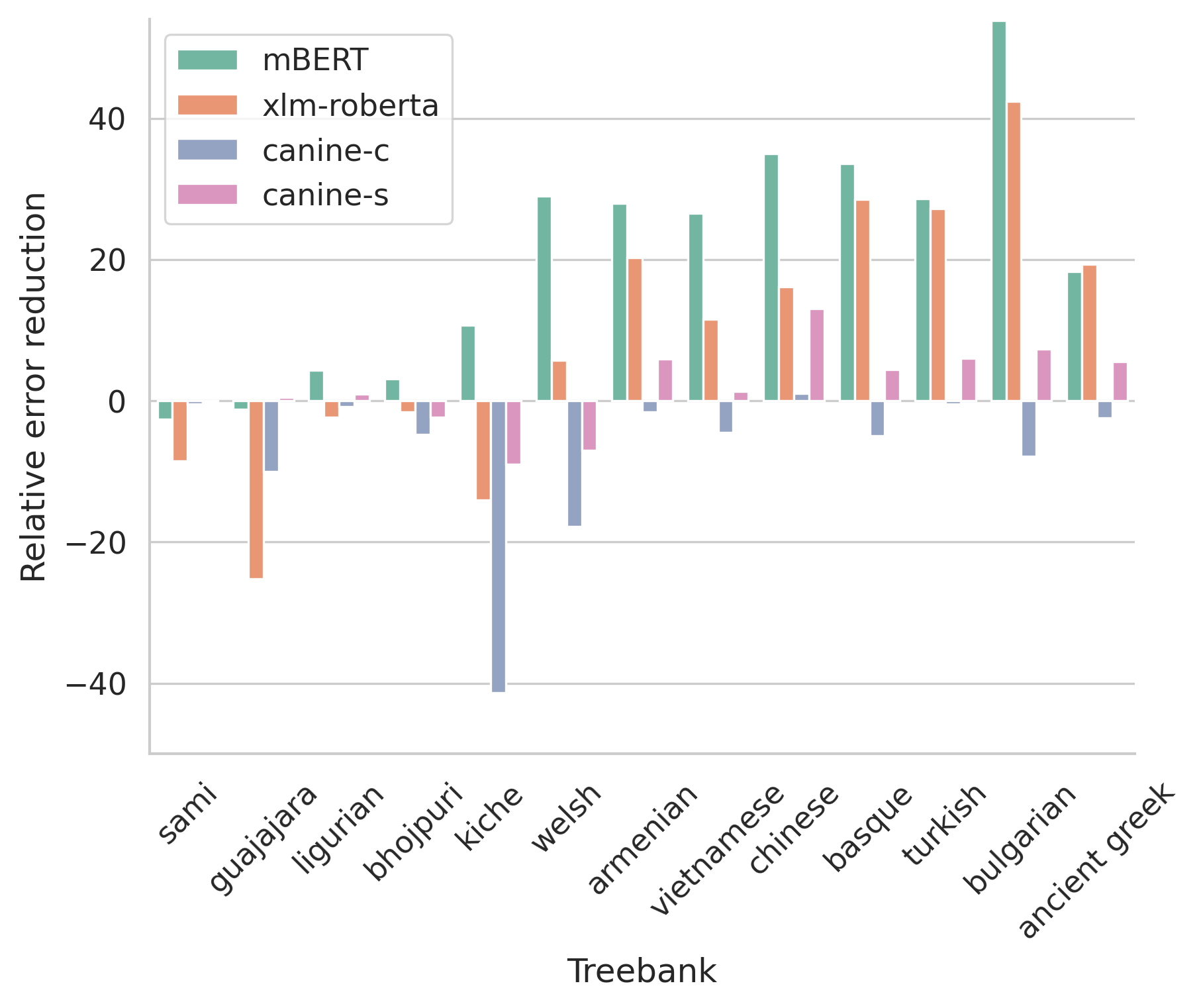}
    \caption{
    $\epsilon_{LAS}$(\texttt{rnd},\texttt{frz}) on the dependency treebanks test sets
    for the \texttt{2p\textsuperscript{b}} encoding.
    }   
    \label{fig:dep_2pb}
\end{figure}

For a clearer picture, Figure \ref{fig:dep_2pb} shows the relative LAS error reductions $\epsilon_{LAS}$(\texttt{rnd},\texttt{frz}) for the 2-planar encoding and sorted by the size of the training set used for the probe. Next, we focus on \texttt{2p\textsuperscript{b}} as previous work has demonstrated its robustness across various configurations \cite{munoz-ortiz-etal-2021-linearizations,strzyz-etal-2019-viable, strzyz-etal-2020-bracketing}.\footnote{The trends for the other encodings are similar and they can be seen in Appendix \ref{app:error_reduction}.} For larger treebanks, whose languages are supported by LLMs, the error reductions between the \texttt{frz} and \texttt{rnd} setups are large, showing that the LLMs encode to some extent dependency structures in their representational space. For languages that are not supported by the LLMs, the error reductions are clearly smaller. This happens for low-resource treebanks, in which only \texttt{mBERT} is able to obtain improvements over the \texttt{rnd} baseline, but also for high-resource ones, such as Ancient Greek (the largest tested treebank), suggesting that the treebank size is not a key factor for the probes (we discuss this in detail \S \ref{section-discussion}).

\paragraph{Frozen (\texttt{frz}) vs fine-tuned (\texttt{ftd}) setup}
Table \ref{tab:dep_finetuned} shows the scores for the fine-tuned models. In this case, \texttt{xlm-roberta} sequence labeling parsers obtain a larger average error reduction, while \texttt{mbert} obtains slightly better results for the \texttt{ftd} setup. The results show that even if under the \texttt{frz} setup dependency structures can be recovered, fine-tuning the whole architecture gives significant improvements. Also, the performance across the board for the fine-tuned models is very competitive for all treebanks supported by the LLMs. Note that even if such results lag below the state of the art (not the target of our work), we rely exclusively on multilingual pretraining vectors, without any powerful parser decoder, such as \citet{kitaev-klein-2018-constituency} for constituent parsing, or \citet{dozat-etal-2017-stanfords} for dependencies.

\paragraph{Encoding comparison} Results from Table \ref{tab:dep_results} show that the three encodings are able to recover a similar amount of syntax. It is worth noting that, although \texttt{r\textsuperscript{h}} performs better for the \texttt{rnd} setup, this does not translate into a better recovering from \texttt{frz} representations. It seems also that \texttt{2p\textsuperscript{b}} recovers more syntactic information in higher-resourced setups (i.e. Bulgarian), while \texttt{r\textsuperscript{h}} and \texttt{ah\textsuperscript{tb}} perform better in lower-resourced configurations (i.e Skolt Sami, Ligurian).

\begin{figure}[htpb!]
    \centering
    \begin{subfigure}[b]{0.23\textwidth}
        \includegraphics[width=\textwidth]{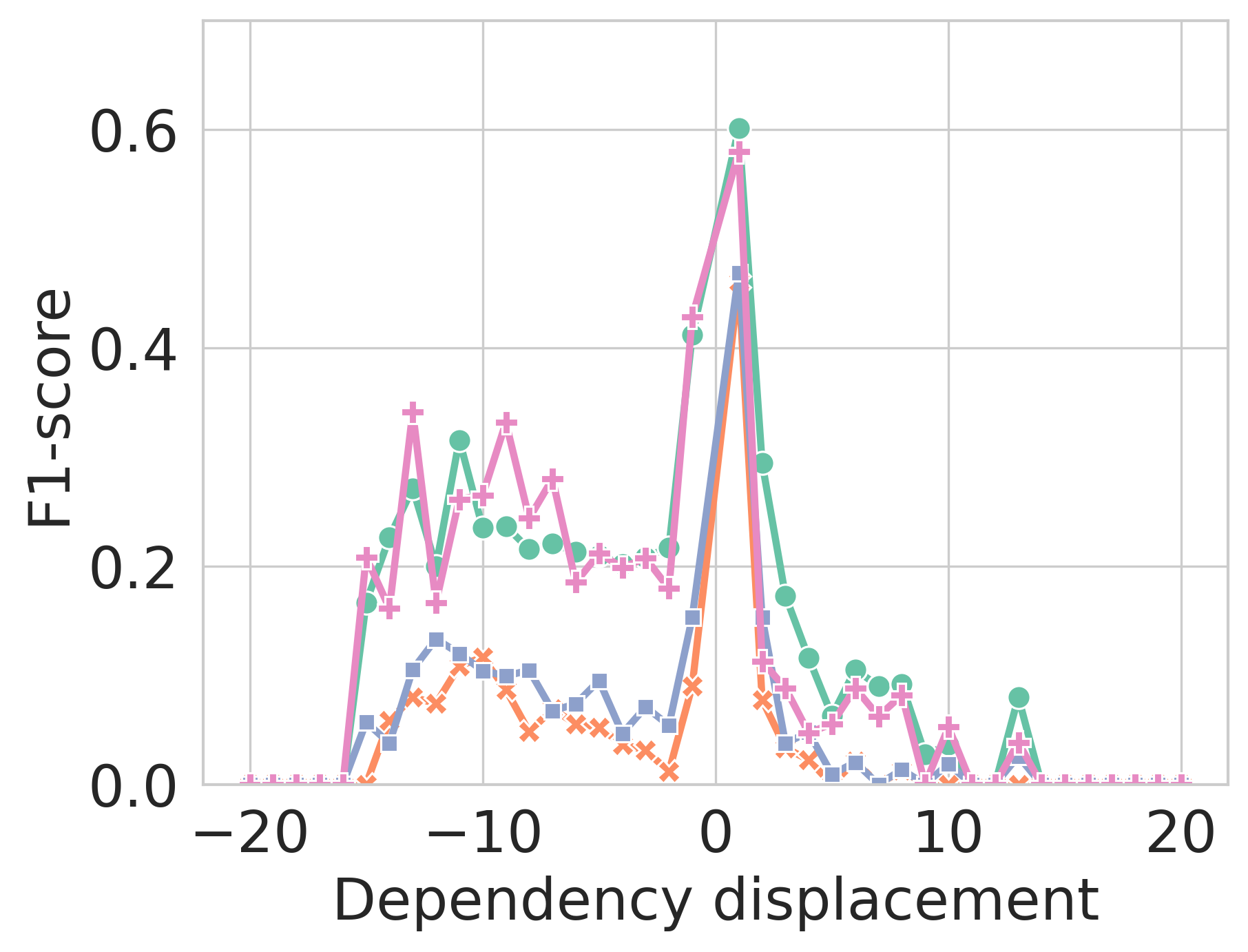}
        \caption{Turkish\textsubscript{BOUN}}
    \end{subfigure}
    ~
    \begin{subfigure}[b]{0.23\textwidth}
        \includegraphics[width=\textwidth]{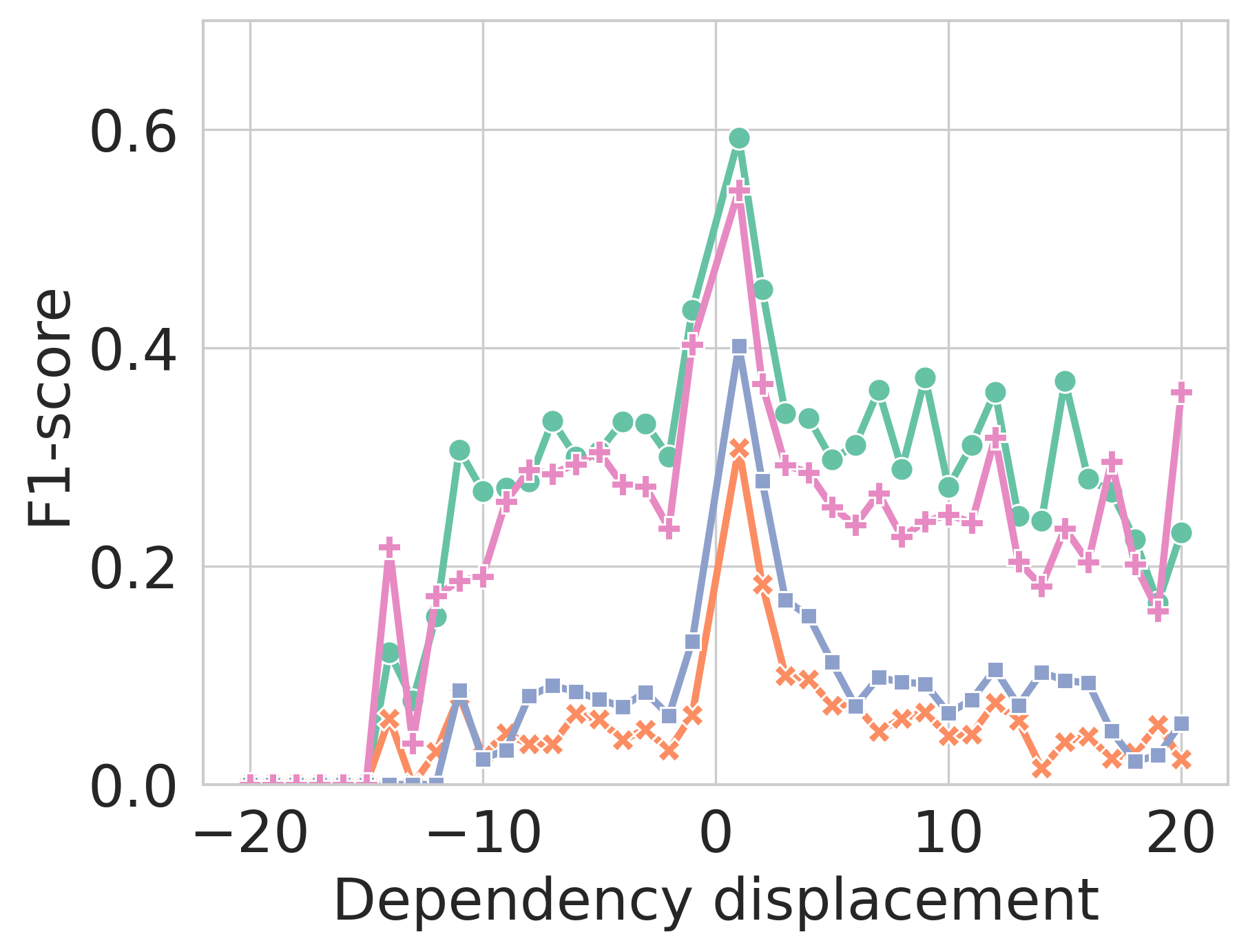}
        \caption{Basque\textsubscript{BDT}}
    \end{subfigure}
    
    \begin{subfigure}[b]{0.23\textwidth}
        \includegraphics[width=\textwidth]{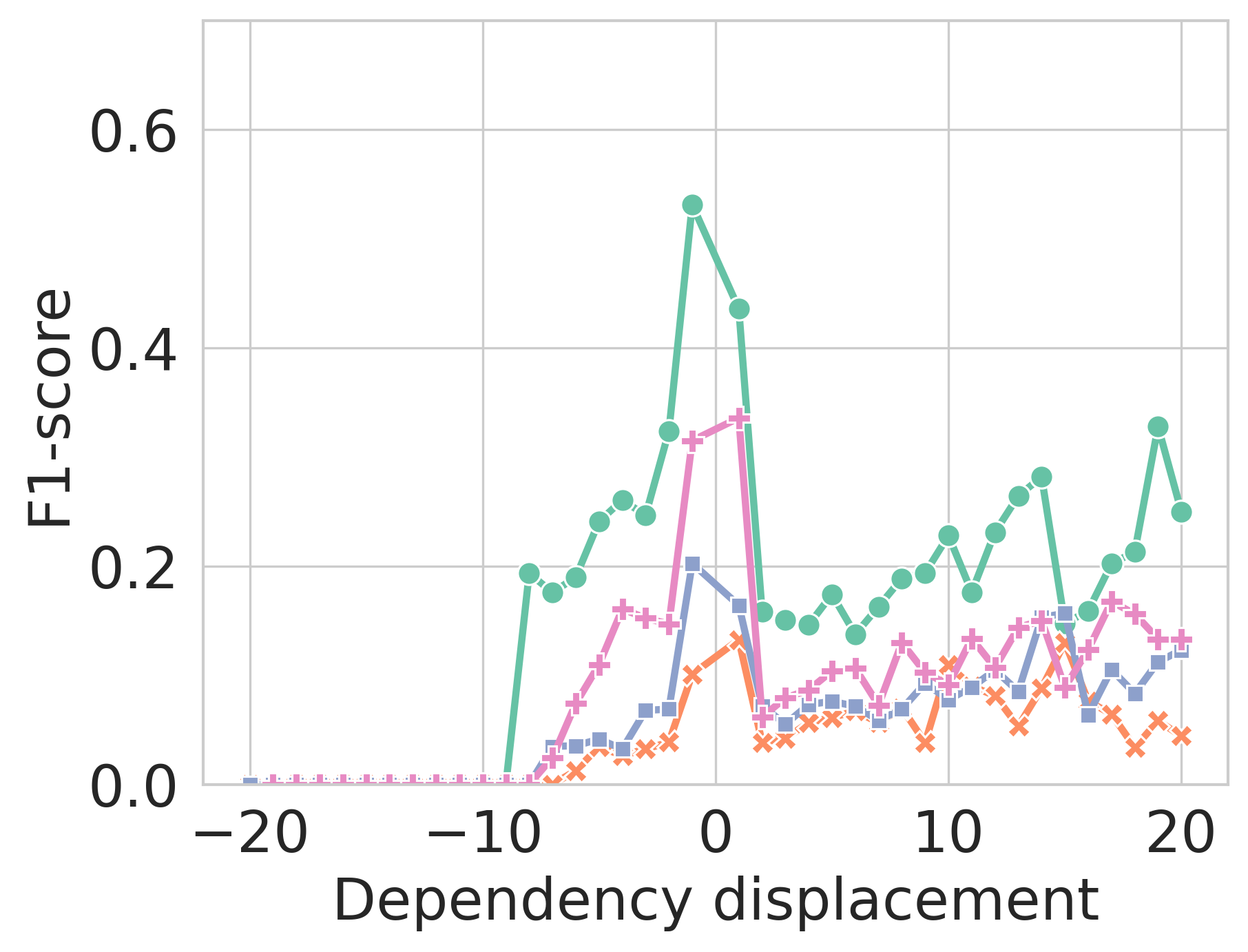}
        \caption{Vietnamese\textsubscript{VTB}}
    \end{subfigure}
    ~
    \begin{subfigure}[b]{0.23\textwidth}
        \includegraphics[width=\textwidth]{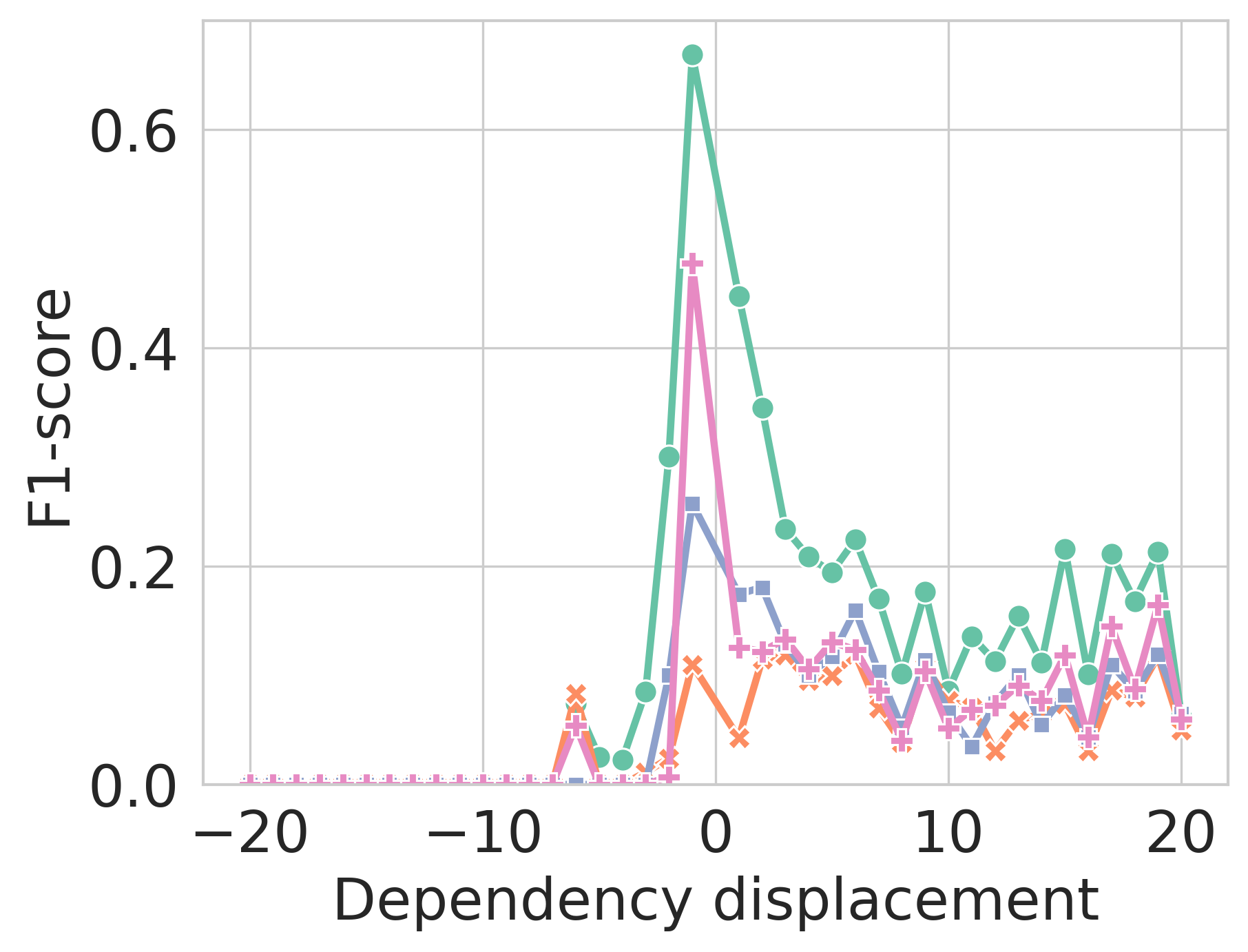}
        \caption{Welsh\textsubscript{CCG}}
    \end{subfigure}
    \begin{subfigure}{0.46\textwidth}
        \includegraphics[width=\textwidth]{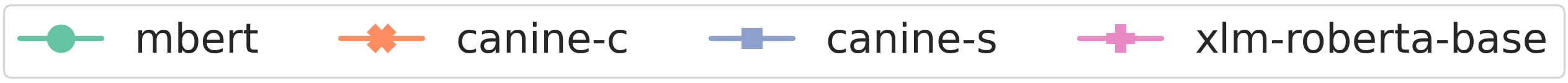}
    \end{subfigure}
    \caption{Average F1 score using \texttt{2p\textsuperscript{b}} for different dependency displacements (signed lengths) and LLMs. We removed displacements occurring less than 10 times. }
    \label{fig:displacements}
\end{figure}

\paragraph{Dependency displacements} Figure \ref{fig:displacements} shows the performance across arcs of different length and direction for the \texttt{frz} models with the \texttt{2p\textsuperscript{b}} encoding over 4 languages: the one with most left arcs (Turkish), with most right arcs\footnote{Guajajara is excluded due to dataset size limitations.} (Vietnamese), and two balanced ones (Basque and Welsh). The multilingual LLMs capture the particularities of languages (for the case of the Welsh\textsubscript{CCG} treebank, even if it is balanced in terms of the number of left/right arcs, left arcs are on average of a distance of 1.6\textsubscript{$\pm$1.8} units while right arcs are of 3.9\textsubscript{$\pm$4.9} units). Also, the LLMs keep the trends across displacements, i.e., no LLM notably changes their expected performance with respect to the others for a specific subset of dependencies.

\begin{table}[thbp!]
    \centering
    \scriptsize
    \begin{tabular}{l|ccc|ccc|ccc|ccc}
    \toprule
    \multirow{2}{*}{Treebank}& \multicolumn{3}{c|}{\texttt{mBERT}} & \multicolumn{3}{c|}{\texttt{xlm-roberta}} & \multicolumn{3}{c|}{\texttt{canine-c}} & \multicolumn{3}{c}{\texttt{canine-s}}\\
     & \texttt{frz} & \texttt{ftd} & \texttt{err} &
     \texttt{frz} & \texttt{ftd} & \texttt{err} &
     \texttt{frz} & \texttt{ftd} & \texttt{err} &
     \texttt{frz} & \texttt{ftd} & \texttt{err}\\
     \midrule
     \textit{Skolt Sami} & 9.2 & 14.6 & 5.9 & 6.9 & 11.2 & 4.6 &  7.2 & 2.6 & -5.0 & 10.3 & 6.3 & -4.5\\
     \textit{Guajajara} & 30.9 & 46.6 & 22.6 & 19.0 & 39.0 & 24.7 & 22.2 & 16.6 & -7.2 & 29.8 & 29.9 & 0.1\\
     \textit{Ligurian} & 7.2 & 28.4 & 22.8 & 1.6 & 24.7 & 23.5 & 3.6 & 0.5 & -3.2 & 5.7 & 0.8 & -5.2 \\
     \textit{Bhojpuri} & 17.0 & 24.9 & 9.5 & 11.8 & 15.0 & 3.6 & 9.1 & 7.4 & -1.9 & 11.3 & 14.2 & 3.6 \\
     \textit{Kiche} & 51.0 & 69.2 & 37.1 & 33.4 & 61.8 & 42.6 & 25.7 & 22.8 & -3.9 & 43.1 & 51.1 & 13.9 \\
     Welsh & 44.9 & 68.9 & 43.6 & 27.9 & 69.0 & 57.0 & 12.5 & 8.0 & -5.1 & 20.5 & 31.5 & 13.8\\
     Armenian & 38.8 & 71.7 & 53.8 & 31.0 & 74.5 & 63.1 & 13.8 & 16.7 & 3.2 & 19.9 & 31.9 & 15.0\\
     Vietnamese &  37.4 & 58.5 & 33.7 & 24.6 & 60.8 & 48.0 & 10.0 & 7.7 & -2.6 & 15.3 & 24.4 & 10.7 \\
     Chinese & 42.1 & 76.2 & 58.9 & 17.6 & 75.9 & 70.7 & 15.6 & 20.0 & 6.3 & 24.9 & 50.4 & 34.0 \\
     Basque & 45.5 & 77.4 & 58.8 & 40.9 & 79.3 & 65.0 & 14.8 & 32.9 & 21.2 & 22.4 & 31.9 & 31.6\\
     Turkish & 42.9 & 68.8 & 45.4 & 41.3 & 48.7 & 53.8 & 18.5 & 37.1 & 22.8 & 23.7 & 48.7 & 32.8 \\
     Bulgarian & 63.4 & 90.4 & 74.0 & 55.3 & 88.1 & 82.1 & 18.0 & 54.1 & 44.0 & 29.4 & 67.0 & 53.3 \\
     \textit{A. Greek} & 23.7 & 51.7 & 36.7 & 23.7 & 67.7 & 57.7 & 11.3 & 37.1 & 29.0 & 18.4 & 45.8 & 33.6\\
     \midrule
     Average & 34.9 & 57.5 & 38.7 & 25.7 & 55.1 & 45.9 & 14.0 & 20.3 & 7.5 & 21.1 & 34.4 & 17.9 \\
    \bottomrule
    \end{tabular}
    \caption{LAS for the \texttt{frz} and \texttt{ftd} setups on the test sets, together with $\epsilon_{LAS}$(\texttt{frz},\texttt{ftd}) 
    for the \texttt{2pb\textsuperscript{b}} encoding for all treebanks and LLMs tested. Languages in italics are absent in the pretraining data of the LLMs.}
    \label{tab:dep_finetuned}
\end{table}

\subsection{Constituent parsing results}\label{section-results-const}

We break down the results comparing frozen \emph{vs}: (i) random, and (ii) fine-tuned weights.

\paragraph{Frozen (\texttt{frz}) \emph{vs} random weights (\texttt{rnd}) setups} Table \ref{tab:const_results} shows the bracketing F1 score across treebanks and the encodings for the two setups. The trend from dependency parsing remains: \texttt{mBERT} outperforms \texttt{xlm-roberta} for all languages, while \texttt{canine-s} outperforms \texttt{canine-c}. In this case, \texttt{canine-s} 
improves over the random baseline for all treebanks, while \texttt{canine-c} only outperforms the random baseline for 3 out of 10 models, which suggests the difficulties that these character-level language models have to model syntax, even if they perform well on other downstream tasks. The exceptions are Korean, German and Chinese. Chinese was also an exception in the case of dependency parsing, so an explanation might be that its writing systems encode more information per character than other languages. Chinese characters represent a whole morpheme, being more similar to a subword token, while Korean Hangul encodes a syllable per character, instead of a single sound as alphabets of the other languages tested.

\begin{figure}[thbp!]
    \centering
    \includegraphics[width=0.45\textwidth]{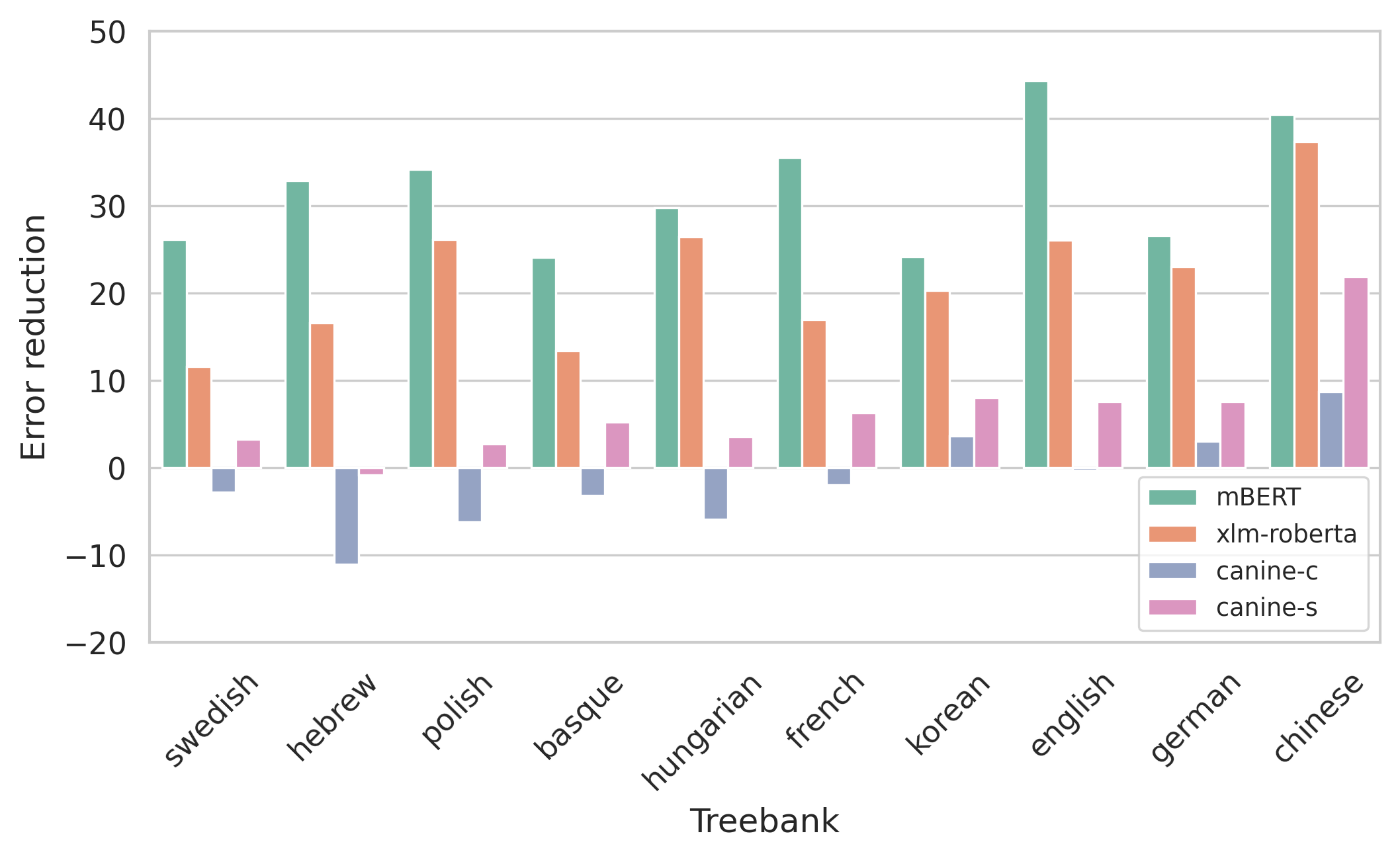}
    \caption{$\epsilon_{F1}$(\texttt{rnd},\texttt{frz}) on the constituent test sets.}
    \label{fig:const_diff}
\end{figure}

Figure \ref{fig:const_diff} shows the error reductions across the board, sorted by the size of the training data used for the probing. In this case, all tested languages are supported by the LLMs, but there are large differences in the size of the training data (e.g., Swedish with 5\,000 sentences \emph{vs} German with 40\,472 sentences). However, we do not see an increase in error reduction when the size of training data grows.

\begin{table}[thbp!]
    \centering
    \scriptsize
    \begin{tabular}{l|gc|gc|gc|gc}
    \toprule
    \multirow{2}{*}{Treebank}& \multicolumn{2}{c|}{\texttt{mBERT}} & \multicolumn{2}{c|}{\texttt{xlm-roberta}} & \multicolumn{2}{c|}{\texttt{canine-c}} & \multicolumn{2}{c}{\texttt{canine-s}}\\
     & \texttt{rnd} & \texttt{frz} & \texttt{rnd} & \texttt{frz} & \texttt{rnd} & \texttt{frz} & \texttt{rnd} & \texttt{frz}\\
     \midrule
     Swedish & 29.8 & 56.0 & 30.1 & 42.3 & 25.5 & 22.7 & 25.5 & 28.7 \\
     Hebrew & 41.5 & 74.5 & 43.3 & 60.0 & 40.8 & 29.8 & 41.0 & 40.2 \\
     Polish & 42.8 & 77.0 & 41.8 & 68.0 & 40.1 & 33.9 & 40.1 & 42.9 \\
     Basque & 32.5 & 56.6 & 33.7 & 47.1 & 36.2 & 33.0 & 36.2 & 41.4 \\
     Hungarian & 40.0 & 69.9 & 39.6 & 66.0 & 37.4 & 31.5 & 37.4 & 41.0 \\
     French & 14.5 & 50.1 & 15.4 & 32.4 & 14.1 & 12.1 & 13.9 & 20.1 \\
     Korean & 33.2 & 57.4 & 32.9 & 53.2 & 33.8 & 37.5 & 33.8 & 42.0 \\
     English & 12.9 & 57.3 & 14.4 & 40.5 & 9.9 & 9.6 & 9.9 & 17.5 \\
     German & 18.9 & 45.4 & 18.1 & 41.2 & 16.4 & 18.5 & 16.4 & 24.0 \\
     Chinese & 16.1 & 56.6 & 8.2 & 45.6 & 16.9 & 25.6 & 17.3 & 39.3 \\
     \midrule
     Average & 28.2 & 60.1 & 27.8 & 49.6 & 27.0 & 25.4 & 27.2 & 33.7 \\
    \bottomrule
    \end{tabular}
    \caption{F-score for the test sets of the constituent treebanks. LLMs analyzed for the \texttt{frz} and \texttt{rnd} setups.}
    \label{tab:const_results}
\end{table}

\paragraph{Frozen (\texttt{frz}) vs fine-tuned (\texttt{ftd)} setup} Table \ref{tab:const_finetuned} compares the bracketing F1-scores for the frozen and  fine-tuned setups, and the behaviors are similar to those obtained in the case of dependency parsing, except for what looks like some empirical outlier, e.g., the fine-tuned \texttt{mBERt} for Hebrew. Hebrew also obtains the lowest error reductions for all LLMs.

\begin{table}[bhtp!]
    \centering
    \scriptsize
    \begin{tabular}{l|ccc|ccc|ccc|ccc}
    \toprule
    \multirow{2}{*}{Treebank}& \multicolumn{3}{c|}{\texttt{mBERT}} & \multicolumn{3}{c|}{\texttt{xlm-roberta}} & \multicolumn{3}{c|}{\texttt{canine-c}} & \multicolumn{3}{c}{\texttt{canine-s}}\\
     & \texttt{frz} & \texttt{ftd} & \texttt{err} &
     \texttt{frz} & \texttt{ftd} & \texttt{err} &
     \texttt{frz} & \texttt{ftd} & \texttt{err} &
     \texttt{frz} & \texttt{ftd} & \texttt{err}\\
     \midrule
     Swedish & 56.0 & 79.4 & 53.2 & 42.3 & 79.9 & 65.2 & 22.7 & 29.6 & 8.9 & 28.7 & 47.8 & 65.2\\
     Hebrew & 74.5 & 75.4 & 3.5 & 59.9 & 76.2 & 40.6 & 29.8 & 36.4 & 9.4 & 40.1 & 53.8 & 22.9 \\
     Polish & 77.0 & 93.4 & 70.9 & 68.0 & 94.0 & 81.2 & 33.9 & 56.6 & 34.3 & 42.9 & 73.9 & 54.3\\
     Basque & 56.6 & 85.0 & 65.4 & 47.1 & 85.1 & 71.6 & 33.0 & 49.1 &  24.0 & 41.4 & 62.7 & 36.2 \\
     Hungarian & 69.8 & 91.5 &71.9 & 66.0 & 92.1 & 76.8 & 31.5 & 52.7 & 30.9 & 41.0 & 65.8 & 42.0\\
     French & 50.1 & 82.2 & 64.3 & 32.4 & 82.6 & 74.3 & 12.1 & 70.2 & 66.1 & 20.1 & 76.0 &70.0 \\
     Korean & 57.4 & 86.4 & 68.1 & 53.2 & 88.0 & 74.4 & 37.5 & 66.0 & 45.6 & 41.9 & 71.4 & 50.8\\
     English & 57.2 & 91.9 & 81.1 & 40.5 & 92.8 & 87.9 & 9.6 & 82.4 & 80.5 & 17.5 & 86.6 & 83.8 \\
     German & 45.4 & 87.3 & 76.7 & 41.1 & 88.4 & 80.3 & 18.5 & 71.7 &65.3 & 24.0 & 77.3 & 70.1 \\
     Chinese & 56.6 & 85.5 & 66.6 & 45.6 & 88.9 & 79.6 & 25.6 & 67.0 & 55.6 & 39.3 & 74.4 & 57.8 \\ 
     \midrule
     Average & 60.1 & 85.8 & 62.2 & 49.6 & 86.8 & 73.2 & 25.4 & 58.2 & 42.1 & 33.7 & 69.0 & 55.3 \\
    \bottomrule
    \end{tabular}
    \caption{F-score for the test sets of the constituent treebanks, LLMs analyzed for the \texttt{ftd} \emph{vs} the \texttt{frz} setup.}
    \label{tab:const_finetuned}
\end{table}

\paragraph{Span lengths}

\begin{figure}[htpb!]
    \centering
    \begin{subfigure}[b]{0.23\textwidth}
        \includegraphics[width=\textwidth]{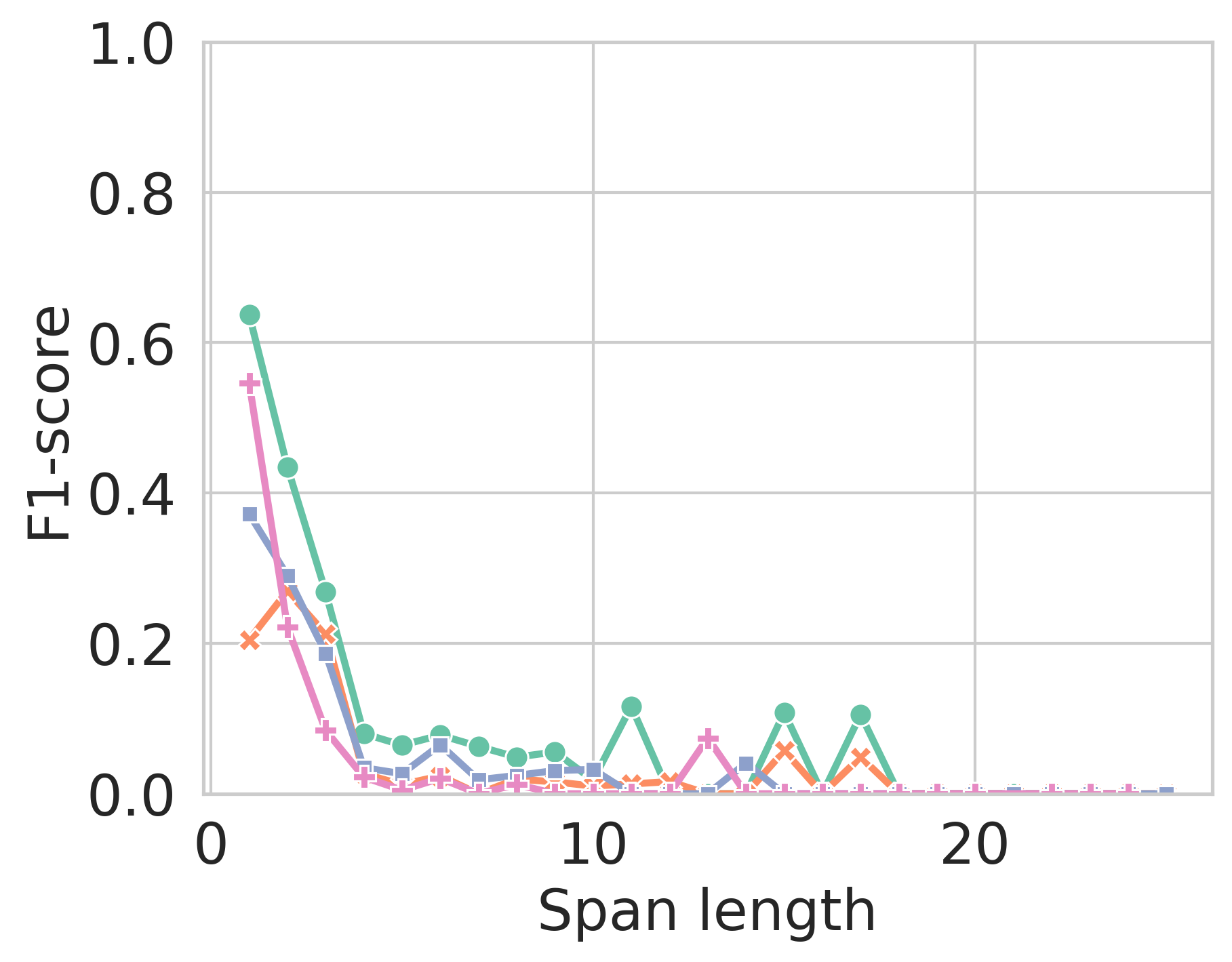}
        \caption{Basque}
    \end{subfigure}
    ~
    \begin{subfigure}[b]{0.23\textwidth}
        \includegraphics[width=\textwidth]{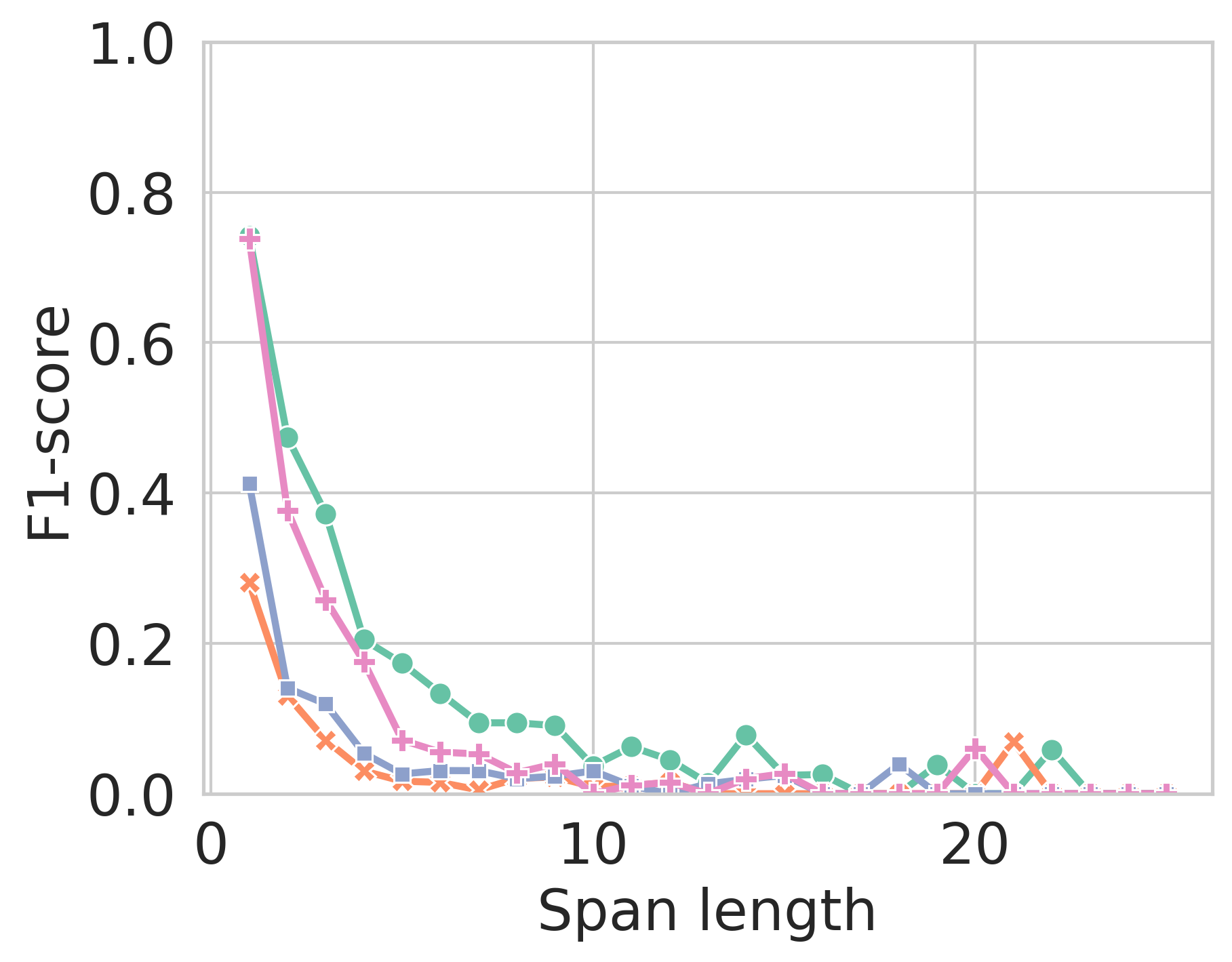}
        \caption{Hungarian}
    \end{subfigure}
    
    \begin{subfigure}[b]{0.23\textwidth}
        \includegraphics[width=\textwidth]{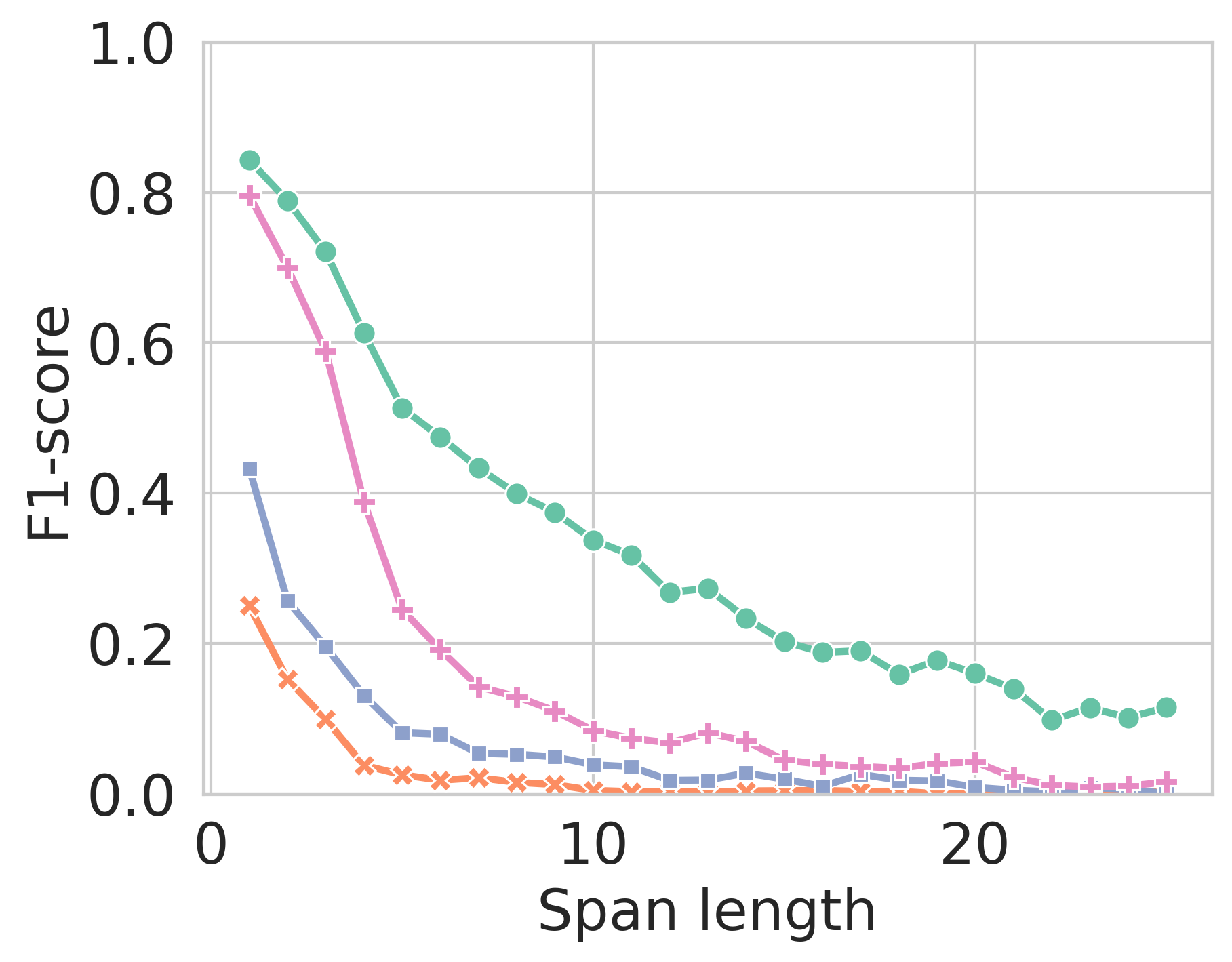}
        \caption{English}
    \end{subfigure}
    ~
    \begin{subfigure}[b]{0.23\textwidth}
        \includegraphics[width=\textwidth]{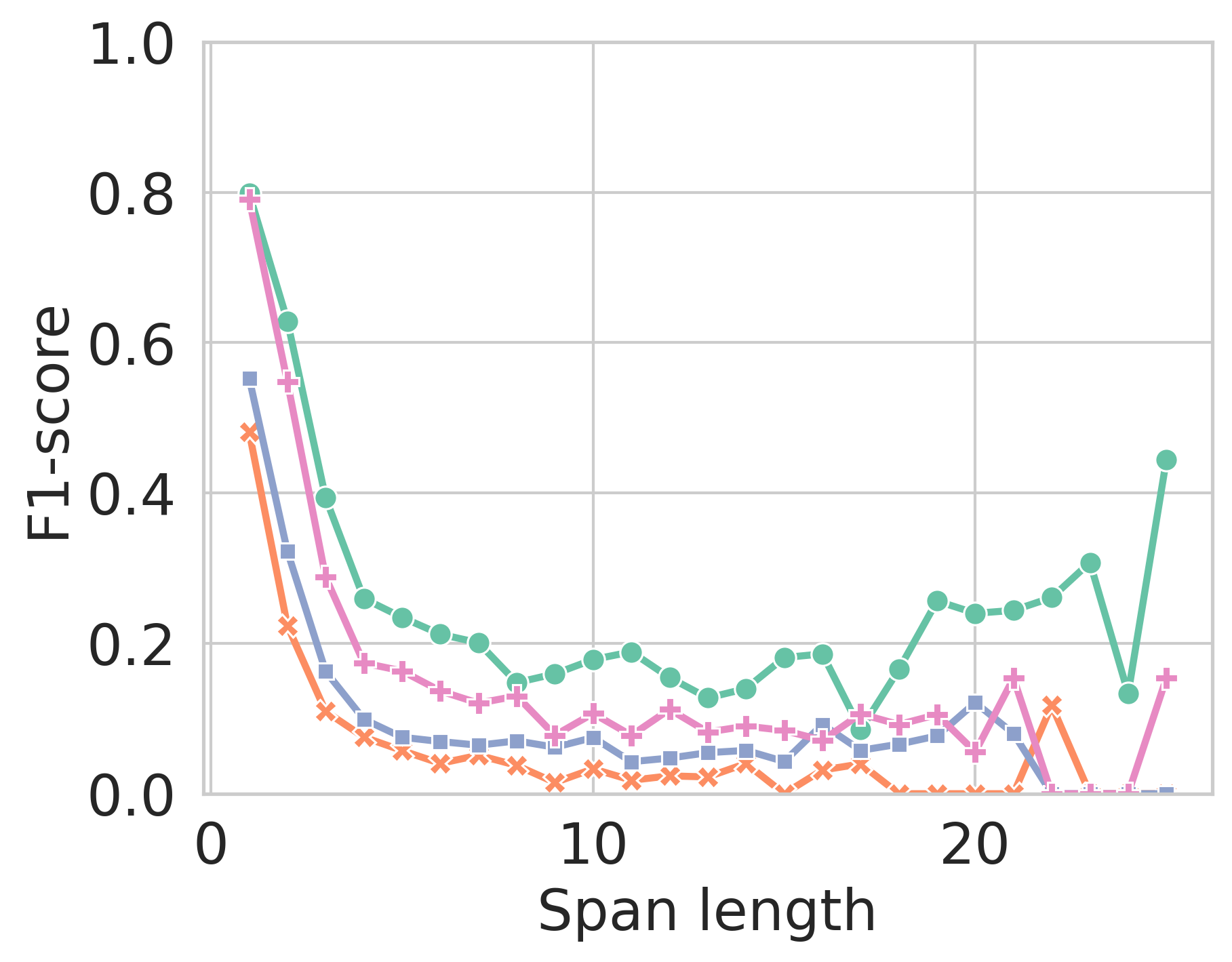}
        \caption{Korean}
    \end{subfigure}
    \begin{subfigure}{0.46\textwidth}
        \includegraphics[width=\textwidth]{plots/legend}
    \end{subfigure}
    \caption{Average F1 score for different span lengths and LLMs.}
    \label{fig:span-lengths}
\end{figure}

Plotting the F1-score for each span length is the rough alternative to dependency displacements in the context of constituent parsing. In Figure \ref{fig:span-lengths} we again show specific examples for some of the studied languages: the most left-branching language (Korean), two balanced ones (Basque and Hungarian), and the most right-branching one (English). Similarly to the case of dependency parsers, the trends across models persist across different span lengths.
They show that, regarding LLMs, \texttt{mbert} obtains the highest F-score for longer spans, while \texttt{xlm-roberta} shows great differences between shorter and longer spans. The \texttt{canine} models perform worse for all lengths. 

\subsection{Discussion}\label{section-discussion}

We now discuss the main insights and potential limitations of the proposed assessment framework.

\paragraph{Pretraining data \emph{versus} Assessment data} An interesting question that arises from multilingual recovery is whether the probe is able to recover the trees due to the size of the training data used for the assessment, although in theory it should be hard to learn by itself by an initially clueless classifier (the random baseline). The experiments show evidence that the size of the training data is not a primary factor to do multilingual, multi-formalism linear probing as sequence labeling. For constituent parsing, we observed that larger treebanks did not come with an increment in the error reductions between the frozen and the random setups, and that the control experiment can thus be used to give an estimate of the amount of structure recoverable from pre-trained representations. Similarly, in the context of dependency parsing, we encountered an analogous situation. Despite the existence of treebanks for languages unsupported by the LLMs, spanning both large (Ancient Greek) and small treebanks (Skolt Sami or Bhojpuri), we observe that treebank size does not significantly impact the reduction in errors between the frozen and random setups.

Either with big or small data, the error reduction between the random and the frozen models is clearly lower than for the treebanks where the language is supported by the LLMs. Among rich-resource treebanks, the size of the data does not have a great influence on the error reductions between the random and frozen weights setups, suggesting that dataset size does not influence the estimates of the dependency structure that are recoverable from the representations.

\paragraph{Language model differences}

The results on the tested LLMs suggest that subword tokenization is necessary to represent syntax, in contrast with token-free models, even if these can later perform well on downstream tasks that require compositionality. Particularly, not only do subword-based models outperform char-based ones, but also \texttt{canine-s}, which is trained using subword loss even though it is a char-level model, performs significantly better than \texttt{canine-c}. It is noteworthy that \texttt{xlm-roberta} generally outperforms \texttt{mBERT} in most downstream tasks, including parsing, as previous studies showed \cite{conneau-etal-2020-unsupervised} and in our fine-tuned results, it performs on par on dependency parsing (Table \ref{tab:dep_finetuned}) and outperforms \texttt{mBERT} in constituency parsing (Table \ref{tab:const_finetuned}). Yet, for the frozen weights setup, \texttt{mBERT}'s representations recovered slightly but consistently better syntactic representations. This suggests that the improvements in how \texttt{xlm-roberta} was trained with respect to \texttt{mBERT}, e.g., training for longer time, or more data, are not key factors to better encode syntax. Additionally, based on our experiments, it appears that \texttt{mBERT} demonstrates a certain level of proficiency in recovering syntax information for the smallest treebanks, particularly for languages not included in the pretraining data (such as Ligurian, Bhojpuri, and Kiche). This suggests a capacity to extend its syntactic knowledge to previously unseen languages, albeit to a limited extent, unlike the other models.

\paragraph{Syntactic formalism}  
Previous studies (e.g., \citet{vilares2020parsing}), hypothesized that pre-trained word vectors might fit better constituent- than dependency-based tasks, since the masked language objective links better with the former formalism, i.e., when a model is learning to unblur a masked token, the constituent structure is to some extent implicit (e.g., an adjective is missing between the determiner and the noun, forming a noun phrase), while dependencies are less obvious. We could not find a clear evidence of this. Although some of the \texttt{frz} models are unable to surpass the \texttt{rnd} baseline in the case of dependencies (while this is not the case for constituents), these instances are languages that are not present in the pretraining data, except for the \texttt{canine} models.

\section{Conclusion}
We proposed a sequence-labeling framework to recover multi-formalism syntactic structures from multilingual LLMs. By mapping syntactic trees to labels we associated output word vectors to labels that encode a portion of the tree, while using a single assessment framework for both constituent and dependency structures. We compared three popular multilingual language models. The results show that subword LLMs can recover a percentage of these structures. 
We evaluated the outcomes by calculating the reduction in errors compared to control models, aiming to gauge the extent to which an LLM can recover specific syntactic structures. The assessment appears reliable and unaffected by variables like the training set's size employed for probing, highlighting that pretraining data is an important factor for recoverability. Last, we found no clear evidence that contextualized vectors encode constituent structures better than dependencies (nor the opposite).

\section*{Limitations}
\paragraph{Physical resources} We did not consider larger language models as we do not have access to the necessary computational resources to run then, hence limiting the scope of our study. We only had access to 2 GeForce RTX 3090, having a total GPU memory of 48 GB, insufficient for fine-tuning many LLMs over different treebanks and formalisms, as in this work.

\paragraph{Language diversity} The constituent treebanks used are all from languages that are relatively rich-resource and are present on the pretraining data of the LLMs. To the best of our knowledge there are no available constituent treebanks from lower-resource languages that are also absent in multilingual LLMs. In consequence, we could not test the effect of absence of pretraining data in order to see if the trends obtained in dependency treebanks prevail here. In addition, for dependency parsing, even a large multilingual resource like Universal Dependencies only has data for about 100 languages, a tiny fraction of the 7\,000 existing human languages.

\paragraph{Interpretation} As mentioned in the introduction, we have to be careful when dealing with probing frameworks. Although we developed solid experiments, and also included control experiments, syntax knowledge is hard to isolate, measure and interpret, so we have tried to be careful with our conclusions.

\section*{Acknowledgments}

We acknowledge the European Research Council (ERC), which has funded this research under the Horizon Europe research and innovation programme (SALSA, grant agreement No 101100615), ERDF/MICINN-AEI (SCANNER-UDC, PID2020-113230RB-C21), Xunta de Galicia (ED431C 2020/11), grant FPI 2021 (PID2020-113230RB-C21) funded by MCIN/AEI/10.13039/501100011033, and Centro de Investigación de Galicia ‘‘CITIC’’, funded by the Xunta de Galicia through the collaboration agreement between the Consellería de Cultura, Educación, Formación Profesional e Universidades and the Galician universities for the reinforcement of the research centres of the Galician University System (CIGUS).

\bibliography{anthology,custom}
\bibliographystyle{acl_natbib}
\clearpage

\appendix

\section{Hyperparameters}\label{app:hyperparameters}
We selected a learning rate of $5\cdot10^{-5}$ for the \texttt{ftd} models and $2\cdot10^{-3}$ for the \texttt{rnd} and \texttt{frz} models based on the results of our preliminary experiments, as the \texttt{ftd} models showed faster convergence. For the three setups, we trained the models during 20 epochs (models had converged at this point). We trained our models on two GeForce RTX 3090 using a batch of 32 on each and a gradient accumulation of 2 for a total batch of 128. Training time of the final models accounts for approximately 60 GPU hours (24 for constituent, 6 per LLM, and 36 for dependency, 8 per LLM).

\subsection{\texttt{mBERT} hyperparameters}
\begin{table}[hbpt!]
    \centering
    \small
    \begin{tabular}{lr}
    \toprule
    Hyperparameter & Value \\
    \midrule
    "attention\textunderscore probs\textunderscore dropout\textunderscore prob" & 0.1\\
    "classifier\textunderscore dropout" & null\\
    "directionality" & "bidi"\\
    "hidden\textunderscore act" & "gelu"\\
    "hidden\textunderscore dropout\textunderscore prob" & 0.1\\
    "hidden\textunderscore size" & 768\\
    "layer\textunderscore norm\textunderscore eps" & 1e-12\\
    "max\textunderscore position\textunderscore embeddings" & 512\\
    "model\textunderscore type" & "bert"\\
    "num\textunderscore attention\textunderscore heads" & 12 \\
    "num\textunderscore hidden\textunderscore layers" & 12\\
    "pad\textunderscore token\textunderscore id" & 0 \\
    "pooler\textunderscore fc\textunderscore size" & 768\\
    "pooler\textunderscore num\textunderscore attention\textunderscore heads" & 12\\
    "pooler\textunderscore num\textunderscore fc\textunderscore layers" & 3 \\
    "pooler\textunderscore size\textunderscore per\textunderscore head" & 128\\
    "pooler\textunderscore type" & "first\textunderscore token\textunderscore transform"\\
    "position\textunderscore embedding\textunderscore type" & "absolute" \\
    "torch\textunderscore dtype" & "float32"\\
    "transformers\textunderscore version" & "4.25.1"\\
    "type\textunderscore vocab\textunderscore size" & 2 \\
    "use\textunderscore cache" & true \\
    "vocab\textunderscore size" & 119547 \\
    \bottomrule
    \end{tabular}
    \caption{Hyperparameters for \texttt{mBERT} models.}
    \label{tab:mbert_params}
\end{table}

 \subsection{\texttt{xlm-roberta-base} hyperparameters}
 \begin{table}[!hbpt]
     \centering
     \small
     \begin{tabular}{lr}
     \toprule
     Hyperparameter & Value \\
     \midrule
     "attention\textunderscore probs\textunderscore dropout\textunderscore prob" & 0.1\\
     "classifier\textunderscore dropout" & null\\
     "eos\textunderscore token\textunderscore id" & 2\\
     "hidden\textunderscore act" & "gelu"\\
     "hidden\textunderscore dropout\textunderscore prob" & 0.1\\
     "hidden\textunderscore size" & 768\\
       "layer\textunderscore norm\textunderscore eps" & 1e-05\\
       "max\textunderscore position\textunderscore embeddings" & 514\\
       "model\textunderscore type" & "xlm-roberta"\\
       "num\textunderscore attention\textunderscore heads" & 12 \\
       "num\textunderscore hidden\textunderscore layers" & 12\\
       "pad\textunderscore token\textunderscore id" & 1 \\
       "position\textunderscore embedding\textunderscore type" & "absolute" \\
       "torch\textunderscore dtype" & "float32"\\
       "transformers\textunderscore version" & "4.25.1"\\
       "type\textunderscore vocab\textunderscore size" & 1 \\
       "use\textunderscore cache" & true \\
       "vocab\textunderscore size" & 250002 \\
     \bottomrule
     \end{tabular}
     \caption{Hyperparameters for \texttt{xml-roberta} models.}
     \label{tab:xlmr_params}
 \end{table}
\newpage
 \subsection{\texttt{canine} hyperparameters}
 \begin{table}[!hbpt]
     \centering
     \small
     \begin{tabular}{lr}
     \toprule
     Hyperparameter & Value \\
     \midrule
     "attention\textunderscore probs\textunderscore dropout\textunderscore prob" & 0.1\\
     "bos\textunderscore token\textunderscore id" & 57344\\
     "downsampling\textunderscore rate" & 4\\
     "eos\textunderscore token\textunderscore id" & 57345\\
     "hidden\textunderscore act" & "gelu"\\
     "hidden\textunderscore dropout\textunderscore prob" & 0.1\\
     "hidden\textunderscore size" & 768\\
       "layer\textunderscore norm\textunderscore eps" & 1e-12\\
       "local\textunderscore transformer\textunderscore stride"& 128\\
       "max\textunderscore position\textunderscore embeddings"& 16384\\
       "model\textunderscore type" & "canine"\\
       "num\textunderscore attention\textunderscore heads" & 12\\
       "num\textunderscore hash\textunderscore buckets"& 16384\\
       "num\textunderscore hash\textunderscore functions" &  8 \\
       "num\textunderscore hidden\textunderscore layers"& 12 \\
       "pad\textunderscore token\textunderscore id"& 0 \\
       "torch\textunderscore dtype" & "float32" \\
       "transformers\textunderscore version" & "4.25.1"\\
       "type\textunderscore vocab\textunderscore size" & 16\\
       "upsampling\textunderscore kernel\textunderscore size" & 4\\
       "use\_cache" & true\\
     \bottomrule
     \end{tabular}
     \caption{Hyperparameters for \texttt{canine-c and -s} models.}
     \label{tab:canine_params}
 \end{table} 
\clearpage
\section{Error reduction for \texttt{r\textsuperscript{h}} and \texttt{ah\textsuperscript{tb}}}\label{app:error_reduction}
\begin{figure}[!hpbt]
    \centering
    \includegraphics[width=0.45\textwidth]{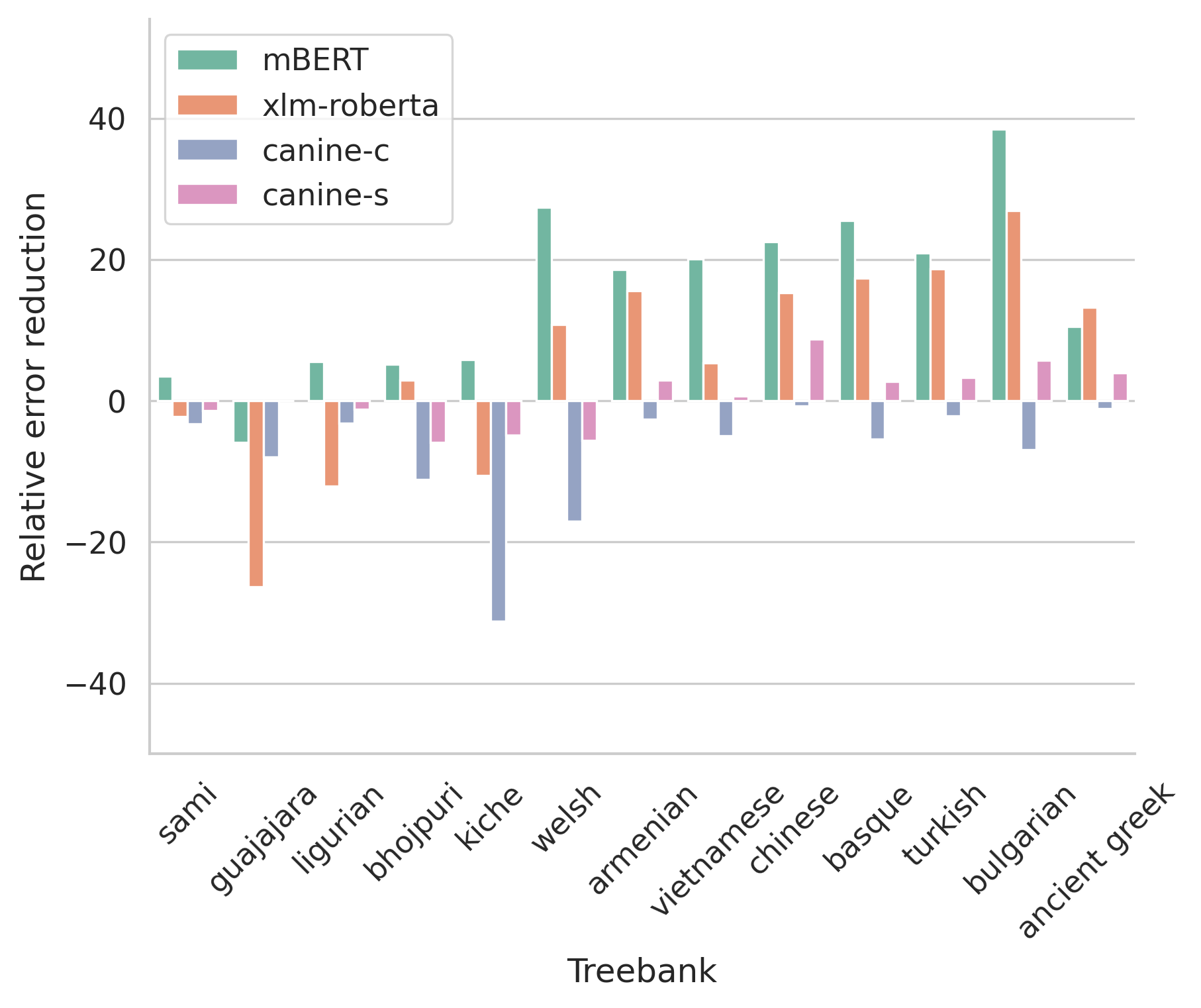}
    \caption{
    $\epsilon_{LAS}$(\texttt{rnd},\texttt{frz}) for the \texttt{r\textsuperscript{h}} encoding for all LLMs tested.
    }
    \label{fig:dep_relative}
\end{figure}

\begin{figure}[hpbt]
    \centering
    \includegraphics[width=0.45\textwidth]{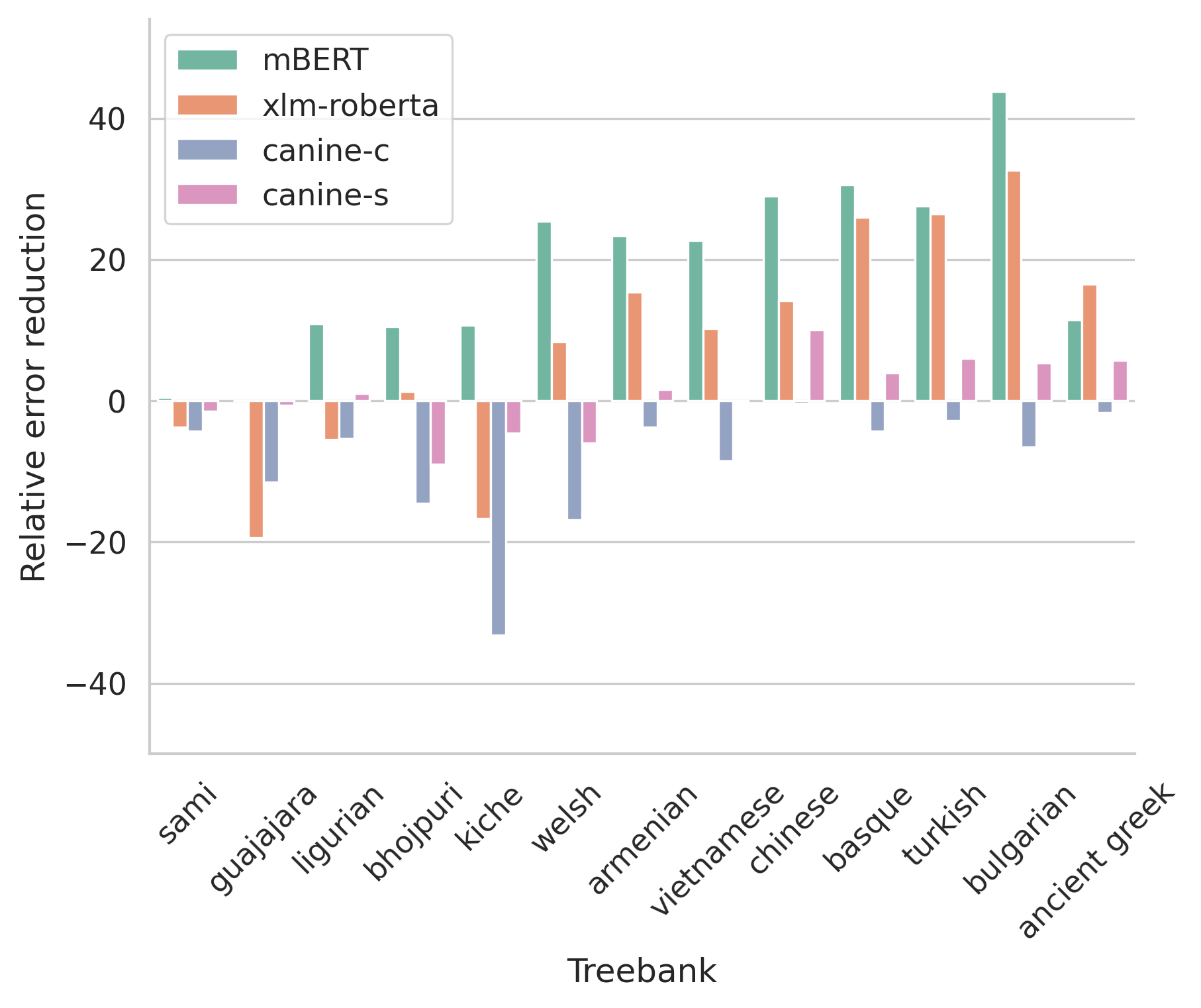}
    \caption{$\epsilon_{LAS}$(\texttt{rnd},\texttt{frz}) for the \texttt{ah\textsuperscript{tb}} encoding for all LLMs tested.}
    \label{fig:dep_archybrid}
\end{figure}

\section{Evaluation scripts}
We used the evaluation scripts \texttt{conll18\_eval.} for dependencies and \texttt{EVALB} for constituencies.
\end{document}